%% file: main.tex
\documentclass[10pt,twocolumn,letterpaper]{article}

\usepackage{cvpr}              
\input{preamble}
\definecolor{cvprblue}{rgb}{0.21,0.49,0.74}
\usepackage[pagebackref,breaklinks,colorlinks,allcolors=cvprblue]{hyperref}
\usepackage{soul}
\usepackage[utf8]{inputenc}
\usepackage{algorithm}
\usepackage{algorithmic}
\usepackage{amsmath}
\usepackage{multirow}
\usepackage{makecell}
\usepackage[table]{xcolor}
\usepackage{booktabs}
\usepackage{xcolor,graphicx}
\usepackage[most]{tcolorbox}
\usepackage{lipsum}
\usepackage{multicol}
\usepackage{lineno}

\def\paperID{6774} 
\def\confName{CVPR}
\def\confYear{2026}

\title{Widget2Code: From Visual Widgets to UI Code via Multimodal LLMs}

\author{Houston H. Zhang$^{1}$, ~ 
Tao Zhang$^{2}$, ~
Baoze Lin$^{1}$, ~
Yuanqi Xue$^{1}$, ~
Yincheng Zhu$^{3}$, ~
Huan Liu$^{1}$, ~ \\
Li Gu$^{4}$, ~
Linfeng Ye$^{2}$, ~
Ziqiang Wang$^{4}$, ~
Xinxin Zuo$^{4}$, ~
Yang Wang$^{4}$, ~
Yuanhao Yu$^{1}$, ~
Zhixiang Chi$^{2\dag}$
\vspace{0.2cm}
\\
${^1}$ McMaster University ~
${^2}$ University of Toronto ~
${^4}$ University of Waterloo ~
${^4}$ Concordia University ~ \\
\href{https://djanghao.github.io/widget2code/}{\textcolor{black}{Project page:} \color[HTML]{ED028C}https://djanghao.github.io/widget2code} \quad $^{\dag}$ Project Lead / Corresponding Author
}

\begin{document}
\maketitle

\input{sec/Abstract}
\input{sec/Intro}

\input{sec/Related_work}
\input{sec/Benchmark}

\input{sec/Method}
\input{sec/Exp}
\input{sec/conclusion}
{
    \small
    \bibliographystyle{ieeenat_fullname}
    \bibliography{cvpr_bib}
}
\definecolor{promptlightblue}{RGB}{230,244,255}
\definecolor{darkblue}{RGB}{0,61,122}
\clearpage

\input{supp_arxiv}


\end{document}

%% file: sec/Abstract.tex
\begin{abstract}
User interface to code (UI2Code) aims to generate executable code that can faithfully reconstruct a given input UI. Prior work focuses largely on web pages and mobile screens, leaving app widgets underexplored. Unlike web or mobile UIs with rich hierarchical context, widgets are compact, context-free micro-interfaces that summarize key information through dense layouts and iconography under strict spatial constraints. Moreover, while (image, code) pairs are widely available for web or mobile UIs, widget designs are proprietary and lack accessible markup. We formalize this setting as the Widget-to-Code (Widget2Code) and introduce an image-only widget benchmark with fine-grained, multi-dimensional evaluation metrics. Benchmarking shows that although generalized multimodal large language models (MLLMs) outperform specialized UI2Code methods, they still produce unreliable and visually inconsistent code. To address these limitations, we develop a baseline that jointly advances perceptual understanding and structured code generation. At the perceptual level, we follow widget design principles to assemble atomic components into complete layouts, equipped with icon retrieval and reusable visualization modules. At the system level, we design an end-to-end infrastructure, WidgetFactory, which includes a framework-agnostic widget-tailored domain-specific language (WidgetDSL) and a compiler that translates it into multiple front-end implementations (e.g., React, HTML/CSS). An adaptive rendering module further refines spatial dimensions to satisfy compactness constraints. Together, these contributions substantially enhance visual fidelity, establishing a strong baseline and unified infrastructure for future Widget2Code research.

\end{abstract}

%% file: sec/Intro.tex
\section{Introduction}
\label{sec:intro}

\begin{figure}[t!]
    \centering
    \begin{subfigure}{0.49\linewidth}
        \centering
        \includegraphics[height=4.9cm]{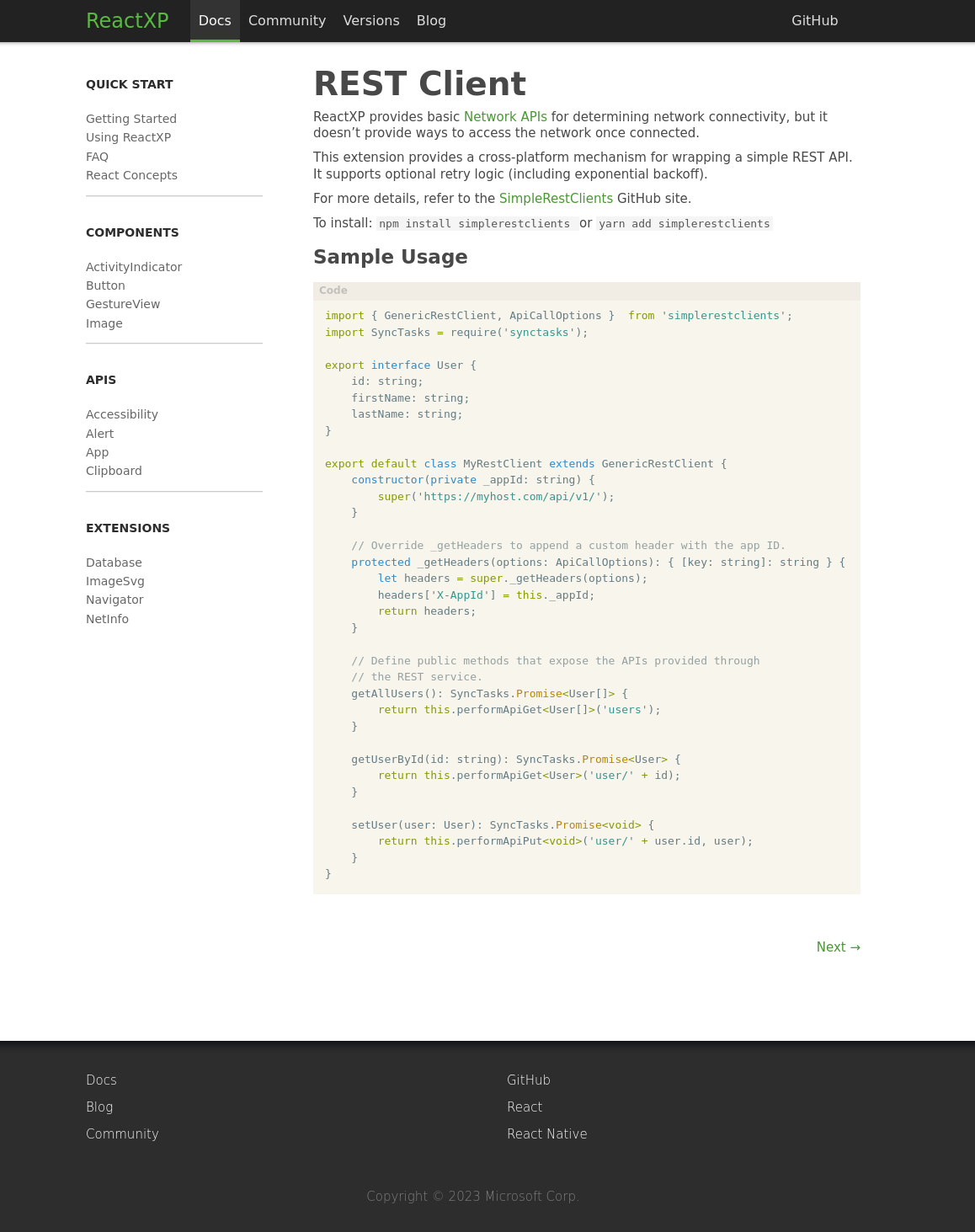}
        \caption{Web UI from Design2Code~\cite{si2025design2code}}
    \end{subfigure}
    \begin{subfigure}{0.49\linewidth}
        \centering
        \includegraphics[height=4.9cm]{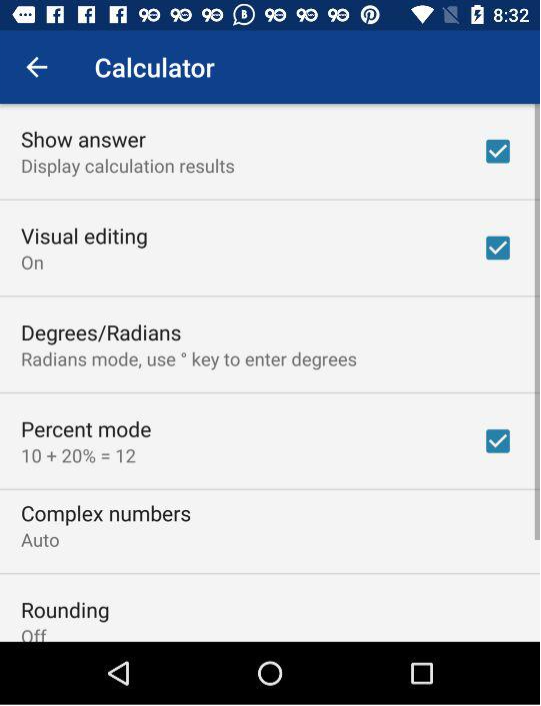}
        \caption{App UI from RICO~\cite{deka2017rico}}
    \end{subfigure}
    \begin{subfigure}{\linewidth}
        \centering
        \includegraphics[width=0.48\linewidth]{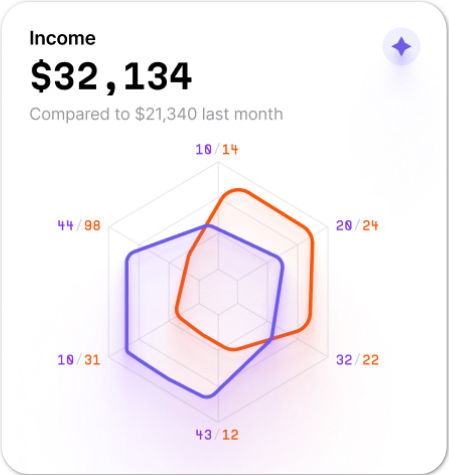}
        \includegraphics[width=0.50\linewidth]{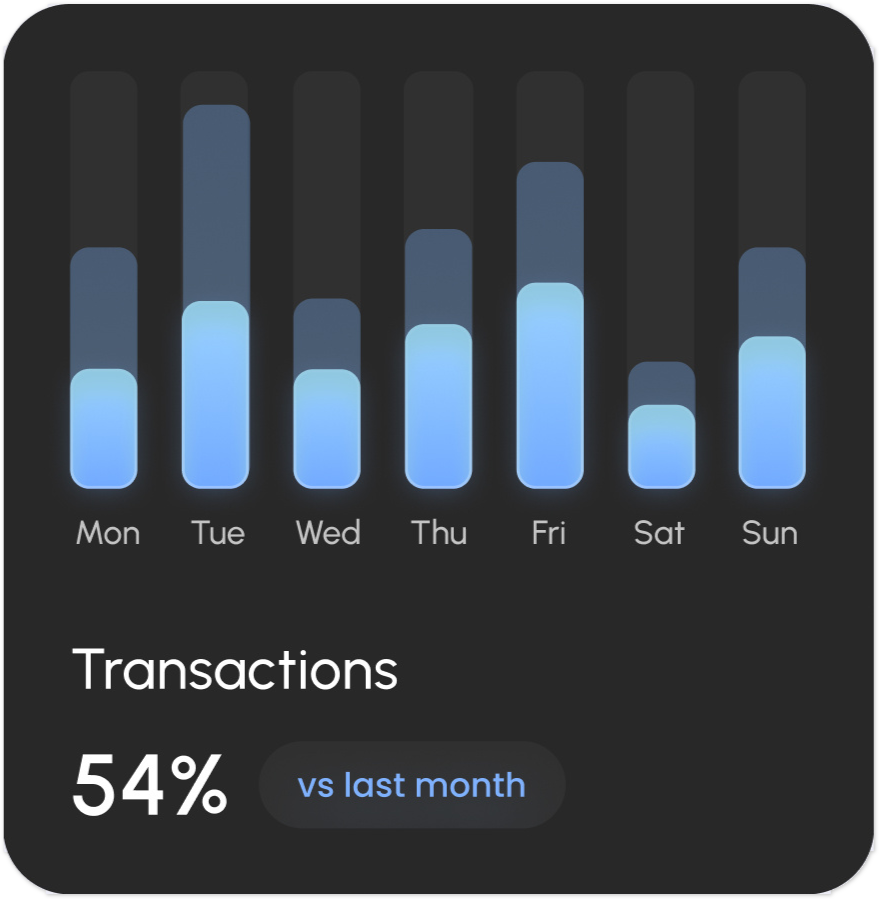}
        \caption{Samples of widgets from our dataset.}
    \end{subfigure}
    \caption{\textbf{Comparison across interface modalities.} Web and mobile UIs provide rich structural and textual context that supports rule-based code mapping, whereas widgets employ dense iconography, embedded graphs, and vivid color schemes within highly constrained layouts. These stylistic and structural compactness factors pose substantial challenges for UI-to-Code reconstruction.}
    \label{fig:front}
\end{figure}

The design of user interface (UI) typically follows a two-stage workflow involving designers and developers ~\cite{beltramelli2018pix2code, nguyen2015reverse}. Designers create visual mockups (layout, style, and interaction intent) using tools such as Figma or Sketch, while developers manually translate these visuals into executable front-end implementations across diverse tech stacks—HTML/CSS, React, Flutter, SwiftUI, and more. This translation requires careful reasoning over layout, hierarchy, and style to ensure visual fidelity across platforms, making it both time-consuming and error-prone~\cite{wan2025divide, nguyen2015reverse, de2020recent}. Automating this process by generating structured, design-faithful code directly from visual designs offers a promising direction to accelerate prototyping, reduce human effort, and improve design consistency~\cite{yang2025ui, jiang2025screencoder}.

With the rapid progress of multimodal large language models (MLLMs)~\cite{yin2024survey, Qwen3VL, hurst2024gpt, comanici2025gemini, ByteDanceSeed16}, UI2Code has shifted from rule-based or supervised pipelines to MLLM-driven generation, which excels at code-related tasks~\cite{wan2025divide, gui2025uicopilot, wu2025mllm, gui2025latcoder, jiang2025screencoder} such as generation~\cite{jiang2025screencoder, wan2025divide, gui2025uicopilot} and repair~\cite{xiao2024interaction2code}. Yet, most efforts focus on web and mobile UIs, where structured supervision is available—web pages naturally come with paired HTML/CSS code~\cite{laurenccon2024unlocking, gui2025webcode2m, si2025design2code}, and mobile UIs can expose view hierarchies, including bounding boxes, layout positions, and component attributes~\cite{deka2017rico}, via the Android Accessibility API. In contrast, widgets are compact, context-free micro-interfaces designed under strict spatial constraints (Fig.~\ref{fig:front}). They lack accessible source code or layout metadata, making faithful reconstruction difficult and leaving widget-to-code largely unexplored. We therefore formalize Widget2Code as a distinct and more challenging visual code generation task, further complicated by the absence of markup, structural annotations, publicly available (image, code) pairs, or even clean widget images.


\begin{figure}[t!]
    \centering
    \includegraphics[width =\linewidth]{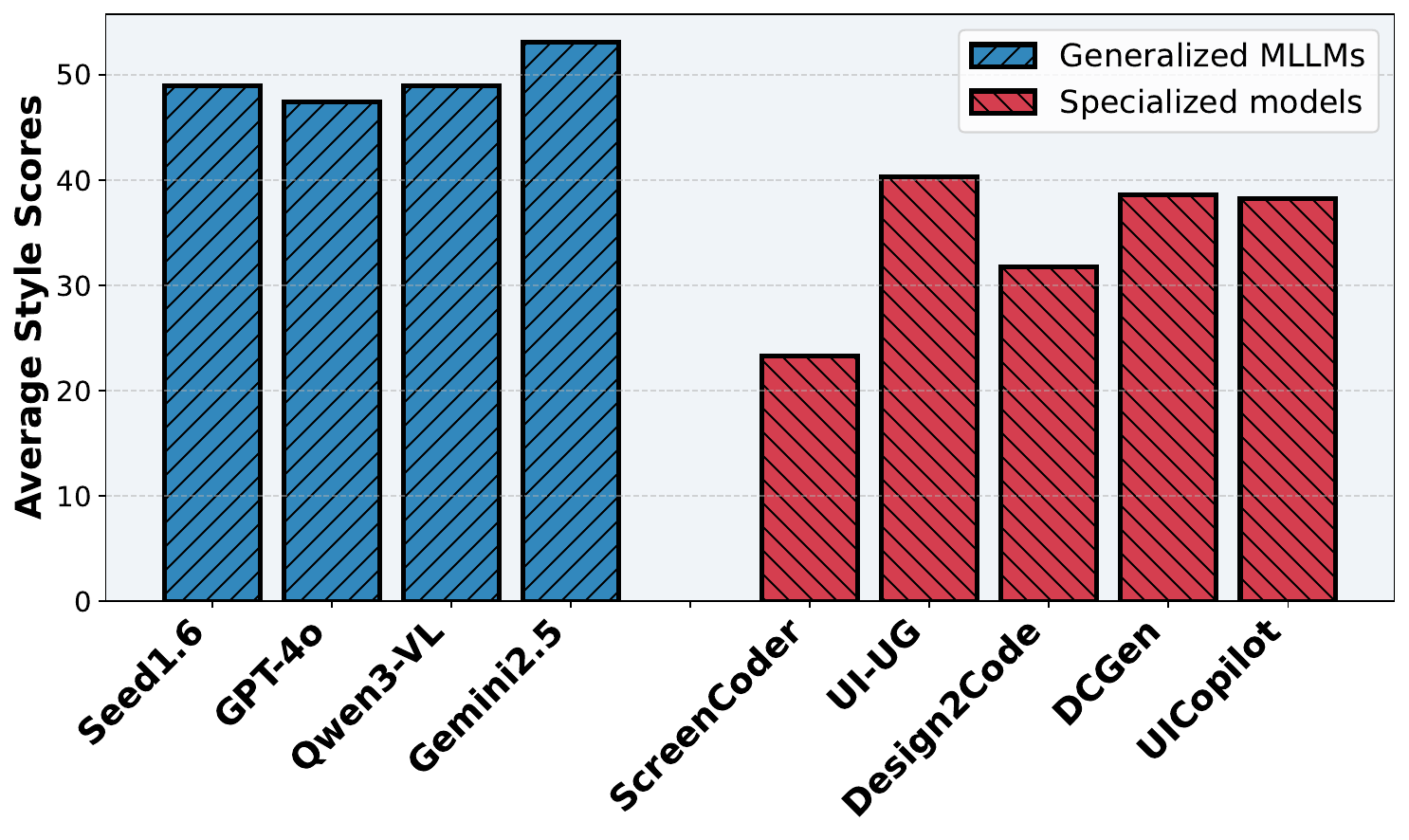} 
    \vspace{-0.7cm}
    \caption{\textbf{Style score comparison on our widget benchmark.} Generalized MLLMs outperform specialized UI2Code models, which are tuned for other UI formats instead of widgets.
}
    \label{fig:motivation}
    \vspace{-0.3cm}
\end{figure}

To investigate this new regime, we first construct the first image-only dataset for Widget2Code by curating and processing widget images from diverse sources. To benchmark existing approaches, we design fine-grained, visual-only evaluation metrics inspired by Apple’s widget design principles, covering five aspects—Layout, Legibility, Perception, Style, and Geometry. Unlike conventional UI2Code metrics that assess generated code or coarse rendered outputs, our metrics capture subtle visual and structural discrepancies that are critical for widgets. 
Benchmarking results show that UI2Code-specialized methods perform worse on widgets than general-purpose MLLMs (e.g., GPT-4o~\cite{hurst2024gpt}; see Fig.~\ref{fig:motivation}), while the latter still produce visually inconsistent and structurally unreliable reconstructions.

We attribute the visual inconsistencies to the intrinsic complexity of widgets, which combine rich iconography, data visualizations (e.g., bar plots, sparklines, pie charts), and artistic styling for visual impact. These elements are challenging to translate directly into UI implementations, often resulting in missing icons, distorted graphics, or a loss of stylistic fidelity. Moreover, existing methods tend to generate verbose code and neglect the rendering stage, causing aspect ratio distortions, dimension mismatches, and occlusions where elements exceed widget boundaries. Together, these issues expose the gap between pixel-level perception and geometry-aware, interpretable code generation, motivating a unified framework that enhances both perceptual understanding and structural controllability.

To establish a strong baseline for future Widget2Code research, we propose a modular framework that advances both perceptual analysis and code generation. 
We first introduce our Perceptual Agent, which follows widget design principles to decompose the input into atomic components and reassemble them into complete layouts, supported by reusable visualization modules. 
At the system level, we present WidgetFactory, an end-to-end infrastructure built around a framework-agnostic domain-specific language (DSL) for widgets and a compiler that maps it to multiple front-end platforms. 
An adaptive rendering module further adjusts spatial dimensions and aspect ratios to preserve compactness and prevent overflow. 
Together, these components form a unified baseline that links visual understanding with widget reconstruction. 
In summary, our contributions are as follows:


\begin{itemize}
    \item We formalize the Widget-to-Code (Widget2Code) task and introduce the first image-only widget dataset with fine-grained, render-level evaluation metrics.
    
    \item We benchmark existing UI2Code models and show that specialized methods underperform general-purpose MLLMs on widgets, exposing key challenges in visual grounding and structural reconstruction.
    
    \item We propose a modular baseline that enhances perceptual understanding through component-level decomposition, icon retrieval, and reusable visualization modules.
    
    \item We develop \textit{WidgetFactory}, an end-to-end infrastructure with a widget-specific DSL, a multi-target compiler, and an adaptive rendering module for geometry-consistent reconstruction and support future research.
\end{itemize}

%% file: sec/Related_work.tex
\section{Related Work}
\label{sec:related_work}

\subsection{Multimodal Large Language Model}
Open-source MLLMs commonly follow a three-module architecture: a visual encoder, an LLM, and a connector that maps encoded visual features to LLM’s embedding space. With high quality datasets~\cite{liu2024improved, chen2024sharegpt4v}, higher and dynamic resolution inputs~\cite{li2024monkey, liu2024llavanext, wang2024qwen2}, stronger connectors~\cite{liu2024improved, li2023blip, alayrac2022flamingo, young2024yi, liu2024infimm, zhu2025connector}, and token efficient techniques~\cite{zhao2025accelerating, ye2025voco, li2024llama, lin2025boosting, chiadapting, chi2025learning}, MLLMs have achieved impressive general-purpose vision-language performance. Recent flagships further extend these capabilities, with multi-level feature fusion in Qwen3-VL~\cite{Qwen3VL}, and native multimodality in Gemini 2.X series~\cite{comanici2025gemini} and GPT-4o~\cite{hurst2024gpt}. However, requiring precise visual grounding and high-fidelity code generation of domain-specific elements such as components, UI to code generation task remains challenging~\cite{jiang2025screencoder, zhou2025declarui}.

\subsection{UI to Code Generation}
To reduce manual effort in GUI prototyping, various approaches for converting visual designs to structured code have been explored. UI2Code methods shifted from computer vision-based element-level layout extraction~\cite{jain2019sketch2code, nguyen2015reverse} and deep learning-based visual and text encoders~\cite{beltramelli2018pix2code, soselia2023learning, xu2021image2emmet} to MLLM-based pipelines. Recent strategies, such as divide-and-conquer prompting~\cite{wan2025divide, wu2025mllm}, segmentation~\cite{chi2025plug} and modeling inter-page relationships~\cite{zhou2025declarui}, test-driven refinements~\cite{wan2025automatically}, and model fine-tuning~\cite{jiang2025screencoder, yang2025ui, laurenccon2024unlocking}, aim to mitigate common issues in MLLMs, including element misclassification, layout misalignment and syntax errors in code synthesis~\cite{jiang2025screencoder, zhou2025declarui}. Despite satisfactory performance in web and phonescreen UIs, current methods either fall short or are not directly applicable to widget-level due to widget’s unique structure and the lack of widget UI-to-code data pairs. Our work follows the recent trend to utilize MLLM, but in distinction from all previous works, we propose an icon retrieval module, and a novel WidgetDSL specialized for widget-level UI structures. To our knowledge, this is the first work in widget-level UI to code generation.

%% file: sec/Benchmark.tex
\section{The Widget2Code Benchmark}


\subsection{Dataset Curation}

Collecting paired (image, code) data for widgets is nearly impossible because widget designs are proprietary and their source code is not publicly available. Clean widget images are also scarce, as no large-scale repositories exist. To address this, we build a crawler to collect widget-related images from platforms such as \textit{Figma}, \textit{Dribbble}, and \textit{Refero}, and we additionally capture screenshots from diverse devices, operating systems, and UI versions.
The collected images often contain multiple or rotated widgets, so we implement an image-processing pipeline (detailed in the Appendix) to detect and extract individual widgets. In total, we gather 12,218 raw samples. After removing duplicates and visually similar items using CLIP features and manual verification, we obtain 2,825 high-quality widgets, with 1,000 used for testing and the rest for development.
Although the test set is modest in size, it is comparable to readily available UI2Code benchmarks such as Design2Code~\cite{si2025design2code} with 485 websites, DeclarUI~\cite{zhou2025declarui} with 250 app pages, and DCGEN~\cite{wan2025divide} with 348 websites, while offering greater visual diversity and a unique focus on the widget domain.




\subsection{Evaluation Metrics}
Prior UI2Code benchmarks primarily evaluate models based on the \textit{generated code} ~\cite{zhou2025declarui, soselia2023learning}, leveraging available ground-truth HTML structures, or rely on \textit{coarse global similarity} metrics such as CLIP score on rendered screenshots. However, these approaches overlook the fine-grained visual fidelity and compositional balance that are particularly critical in widget reconstruction. Design2Code~\cite{si2025design2code} introduces low-level element matching metrics for general UIs, but its evaluation still depends on text bounding boxes parsed from code, making it unsuitable for purely visual widget comparisons. To address this limitation, we propose a set of \textbf{fine-grained, visual-only metrics} that quantitatively assess layout structure, color style, and legibility—drawing inspiration from Apple’s Human Interface Guidelines for Widgets, which emphasize clarity, balance, and visual harmony.

\textbf{Layout.} 
We assess structural fidelity using three complementary metrics aligned with Apple’s widget design principles of balanced composition, proportional scaling, and consistent padding. 
(1) \textit{Margin Symmetry (\textbf{Margin})} measures uneven spacing across margins, indicating misaligned or crowded layouts; 
(2) \textit{Content Aspect Ratio Similarity (\textbf{Content})} captures distortions in the content’s bounding-box proportions that disrupt visual balance; and 
(3) \textit{Area Ratio Similarity (\textbf{Area})} evaluates similarities in the relative area of internal components, reflecting consistencies in visual weight and spacing hierarchy.

\input{sec/Algo_Fig_Tab/maintable}
\begin{figure*}[t]
    \centering
    \includegraphics[width=1\linewidth]{imgs/qualitative_comparison.pdf}
    \caption{\textbf{Qualitative comparison of existing methods on the Widget2Code benchmark.} 
Specialized UI2Code models fail to reproduce the input appearance, while generalized MLLMs generate more coherent results. 
However, all models still exhibit issues such as content overflow, structural misalignment, and color inconsistencies, highlighting the challenges of faithful widget reconstruction. 
}
    \label{fig:comparison}
\end{figure*}
\textbf{Legibility.} 
We evaluate textual clarity and readability through three metrics that reflect Apple’s guideline of maintaining clear, high-contrast typography. 
(1) \textit{Text Jaccard (\textbf{Text})} measures the overlap between OCR-extracted words, capturing semantic consistency in reproduced text; 
(2) \textit{Contrast Similarity (\textbf{Contrast})} quantifies global luminance deviation, ensuring comparable overall readability; and 
(3) \textit{Local Contrast Similarity (\textbf{LocCon})} compares contrast specifically within detected text regions, emphasizing faithful preservation of foreground–background separation. 

\textbf{Style.} 
We measure visual and stylistic fidelity using three metrics that reflect Apple’s emphasis on color harmony, vibrancy, and visual balance. 
(1) \textit{Palette Distance (\textbf{Palette})} evaluates similarity in overall hue distribution, capturing consistency in color themes; 
(2) \textit{Vibrancy Consistency (\textbf{Vibrancy})} assesses alignment in saturation levels, ensuring comparable vividness and tonal intensity; and 
(3) \textit{Polarity Consistency (\textbf{Polarity})} verifies that foreground and background luminance relationships are preserved, maintaining the intended light–dark composition. 

\textbf{Perceptual.} 
We additionally report standard perceptual metrics, \textit{SSIM}~\cite{wang2004image}, \textit{LPIPS}~\cite{zhang2018unreasonable}, and \textit{CLIP Score}, to provide reference points for global visual similarity. 
While these measures effectively capture structural and semantic alignment at the image level, they are less sensitive to the fine-grained geometric and stylistic details critical in widget reconstruction. 
We thus include them primarily for completeness and comparability with prior UI2Code benchmarks.

\textbf{Geometry.} 
We evaluate geometric fidelity by comparing the overall aspect ratio and normalized image dimensions between the input and reconstructed widgets. 
This metric ensures that the generated widget preserves the intended size class and spatial proportions of the original design, preventing visual distortions.

Together, these metrics provide a holistic and quantitative assessment of widget reconstruction quality, complementing coarse global metrics and enabling fine-grained evaluation of multimodal models. We include detailed derivation and implementation in the Appendix.

\subsection{Benchmark Results and Analysis}

We benchmark the two groups of works: 1) generalized MLLMs like GPT-4o~\cite{hurst2024gpt}, Gemini2.5-Pro~\cite{comanici2025gemini}, Seed1.6-Thinking~\cite{ByteDanceSeed16}, Qwen3-VL~\cite{Qwen3VL} and Qwen3-VL-235b. 2) specialized UI2Code methods built based upon MLLMs, e.g., ScreenCoder~\cite{jiang2025screencoder}, UI-UG~\cite{yang2025ui}, DCGen~\cite{wan2025divide}, UICopilot~\cite{gui2025uicopilot}, LatCoder~\cite{gui2025latcoder}, Design2Code~\cite{si2025design2code}, and WebSight-VLM-8B~\cite{laurenccon2024unlocking}. For the MLLMs, we prompt the input widget with the system prompt (provided in the Appendix), asking the model to return the HTML code for reconstruction.

\noindent \textbf{Quantitative results: } Table~\ref{tab:main_benchmark} presents the benchmarking results on our widget test set. 
Specialized UI2Code models, although effective on web and mobile datasets, exhibit pronounced performance degradation on widgets. 
In contrast, general-purpose MLLMs such as GPT-4o and Gemini achieve higher visual fidelity, suggesting better perceptual grounding, yet they still struggle to preserve structural consistency and stylistic accuracy. 
Moreover, all methods fail to reproduce the exact widget dimensions, even when explicitly \textit{prompted to match the input size}. 
These results indicate that existing models lack the inductive bias and representation capacity required for compact and context-free widget reconstruction.

\noindent \textbf{Qualitative results: } Fig.~\ref{fig:comparison} shows the qualitative illustration. The specialized ones are unlikely to reconstruct the input image, while the generalized MLLMs are capable to produce reasonable and better results. Still, we find that they struggle with occlusion, where the content is shown beyond the widget; not able to reconstruct the input layouts and style and produce results with mismatched color, etc.

%% file: sec/Algo_Fig_Tab/maintable.tex
\begin{table*}[!t]
  \tabcolsep3.2pt
  \centering
    \caption{\textbf{Benchmarking results of generalized MLLMs and specialized UI2Code methods on our widget test set using the proposed fine-grained evaluation metrics.} Specialized UI2Code models perform poorly as they are optimized for web or mobile layouts rather than compact widget structures, while generalized MLLMs demonstrate stronger adaptability and overall performance.
  }
    \footnotesize
  \begin{tabular}{c|lccccccccccccc}
    \toprule
    & \multirow{2}{*}{\textbf{Methods}}  & \multicolumn{3}{c}{\textbf{Layout}} & \multicolumn{3}{c}{\textbf{Legibility}} & \multicolumn{3}{c}{\textbf{Style}} & \multicolumn{3}{c}{\textbf{Perceptual}} & \textbf{Geometry}   \\
    \cmidrule(r){3-5} \cmidrule(r){6-8} \cmidrule(r){9-11} \cmidrule(r){12-14} \cmidrule(r){15-15}
    & & \textbf{Margin} & \textbf{Content} & \textbf{Area} & \textbf{Text} & \textbf{Contrast} & \textbf{LocCon} & \textbf{Palette} & \textbf{Vibrancy} & \textbf{Polarity} & \textbf{SSIM} & \textbf{LPIPS}$\downarrow$ & \textbf{CLIP} & \\
    \midrule
    \multirow{5}{*}{\rotatebox{90}{Generalized}} & Seed1.6-Thinking & 65.02 & 59.19 & 74.88 & \underline{62.82} & 62.10 & 55.87 & 47.06 & 44.38 & 55.52 & 0.644 & 0.341 & 0.834 & 94.24 \\
    & Gemini2.5-Pro & 65.35 & \underline{62.74} & \underline{79.66} & 59.48 & \underline{62.64} & 60.52 & \underline{48.99} & \underline{47.77} & \underline{62.54} & 0.701 & \textbf{0.309} & \textbf{0.844} & 90.25\\
    & GPT-4o & 63.48 & 59.04 & 64.41 & 60.20 & 57.10 & 53.65 & 47.03 & 42.08 & 53.15 & 0.698 & 0.338 & 0.780 & 91.93 \\
    & Qwen3-VL & 64.75 & 60.15 & 69.53 & 61.17 & 60.87 & 61.12 & 47.44 & 44.50 & 54.85 & 0.703 & \underline{0.334} & 0.800 & 95.15 \\
    & Qwen3-VL-235b & \underline{66.10} & 60.25 & 69.93 & 61.94 & 60.34 & \underline{61.73} & 47.97 & 45.33 & 54.06 & 0.694 & 0.336 & 0.803 & \underline{96.28} \\
    \midrule
    \multirow{7}{*}{\rotatebox{90}{Specialized}} & Design2Code & 36.34 & 47.81 & 49.68 & 17.50 & 52.90 & 18.31 & 30.92 & 31.89 & 32.54 & 0.512 & 0.494 & 0.610 & 15.72 \\
    & DCGen & 43.17 & 40.14 & 64.55 & 50.36 & 52.13 & 35.02 & 35.56 & 26.21 & 54.03 & 0.598 & 0.400 & 0.753 & 31.59\\ 
    & LatCoder & 41.25 & 43.39 & 76.35 & 48.75 & 56.73 & 34.90 & 41.61 & 36.26 & 49.97 & 0.595 & 0.381 & 0.764 & 28.22 \\
    & UICopilot & 59.20 & 6.95 & 33.94 & 40.94 & 55.19 & 42.56 & 39.22 & 35.08 & 40.51 & \underline{0.709} & 0.354 & 0.691 & 28.40 \\
    & WebSight-VLM-8B & 32.99 & 22.46 & 54.50 & 1.37 & 54.24 & 28.45 & 31.92 & 31.78 & 35.04 & 0.536 & 0.478 & 0.536 & 27.56 \\
    & ScreenCoder & 22.19 & 11.46 & 31.04 & 13.77 & 25.35 & 24.66 & 32.15 & 33.62 & 3.99 & 0.101 & 0.512 & 0.582 & 44.56 \\ 
    & UI-UG & 52.97 & 47.90 & 72.93 & 12.89 & 55.66 & 31.47 & 38.67 & 32.36 & 49.83 & 0.594 & 0.403 & 0.577 & 23.35\\
    \midrule
    \rowcolor{gray!20} \multicolumn{2}{c}{Widget2Code (Ours)} & \textbf{72.15} & \textbf{66.08} & \textbf{82.24} & \textbf{70.60} & \textbf{66.20} & \textbf{64.06} & \textbf{58.09} & \textbf{51.38} & \textbf{63.28} & \textbf{0.721} & 0.335 & \underline{0.838} & \textbf{100.00} \\
  \bottomrule
  \end{tabular}
\label{tab:main_benchmark}
\end{table*}


%% file: sec/Method.tex
\section{Widget2Code Baseline Framework}

\begin{figure}[t!]
    \centering
    \includegraphics[width =0.9\linewidth]{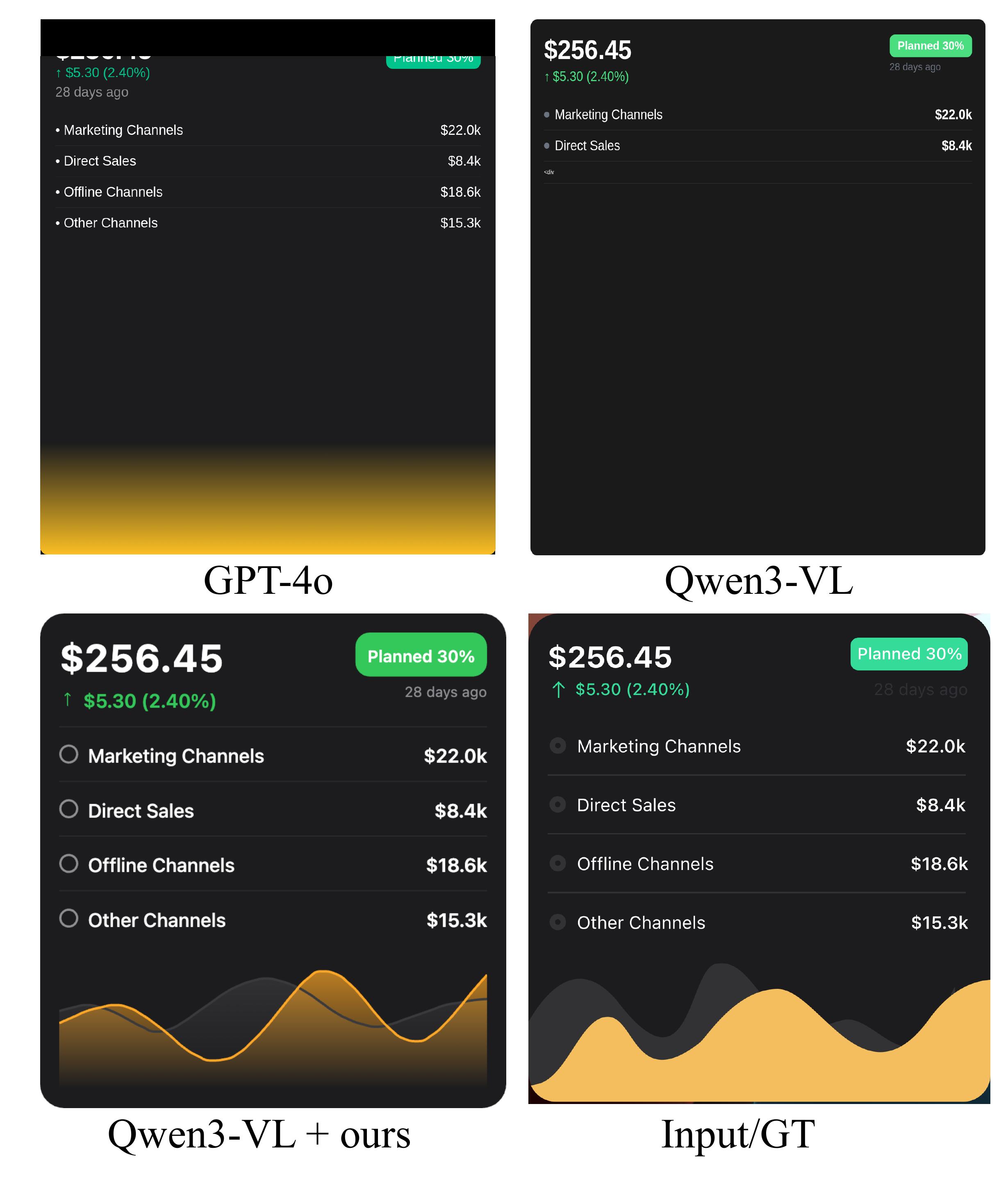} 
    \vspace{-0.3cm}
    \caption{\textbf{A visual example of a challenging case.} 
Failure cases in generalized MLLMs commonly arise from layout misinterpretation, content occlusion, missing components, and difficulty reconstructing complex graphs.
}
    \label{fig:motivation_1}
    \vspace{-0.4cm}
\end{figure}

\noindent \textbf{Motivation.} 
Benchmarking results show that generalized MLLMs outperform specialized UI2Code models on widgets, as the latter are primarily designed for web or mobile interfaces. 
However, even the generalized models remain insufficient, often producing visual inconsistencies between the input and reconstructed widgets. 
Fig.~\ref{fig:motivation_1} illustrates a challenging example where these failures become evident. 
The root cause lies in the inherent complexity of widget design, which combines diverse iconography, compact data visualizations, and stylistic embellishments for visual expressiveness. 
These heterogeneous elements are difficult to translate into executable code from visual inputs alone. 
In addition, the generated code is often verbose and redundant, making the generation process unreliable and hard to control.

\begin{figure*}[t]
    \centering
     \includegraphics[width=1\linewidth]{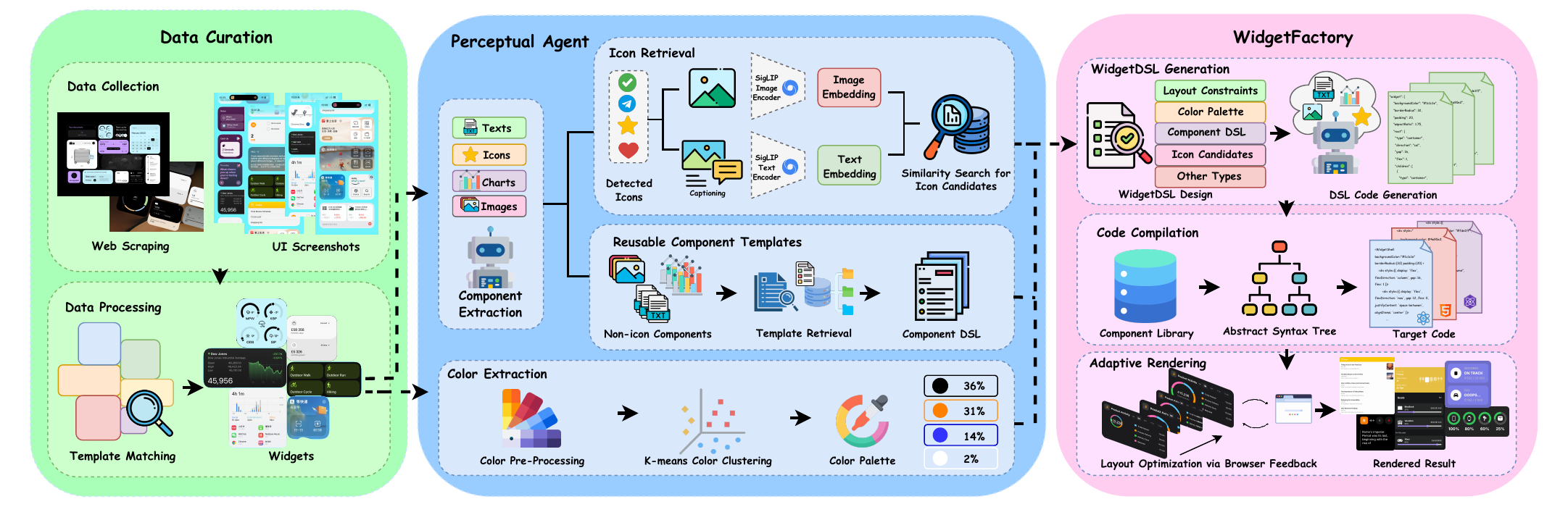}
    \caption{\textbf{Overview of our Widget2Code framework. }
The pipeline consists of three components: 
(1) data curation, which collects and processes widget images to construct our benchmark; 
(2) the Perceptual Agent, which decomposes the input into atomic components and extracts visual, semantic, and stylistic cues; 
and (3) WidgetFactory, an end-to-end infrastructure that generates, compiles, and adaptively renders WidgetDSL to reconstruct the input widget.
}
    \label{fig:overview}
    \vspace{-0.2cm}
\end{figure*}

\noindent \textbf{Baseline Overview.} 
To address these challenges, we propose a concise yet robust baseline that strengthens both perceptual understanding and system-level generation as (Fig.~\ref{fig:overview}). 
At the perceptual level, the framework adheres to widget design principles and formulates a \textit{Perceptual agent} that incrementally assembles layouts from atomic components into complete structures. 
This process integrates icon retrieval, reusable visualization modules to preserve semantic consistency and stylistic fidelity. 
At the system level, we develop an infrastructure termed \textit{WidgetFactory}, which defines a framework-agnostic domain-specific language (WidgetDSL) tailored for widgets. 
A dedicated compiler converts this intermediate WidgetDSL into multiple front-end frameworks, including React and HTML. 
An adaptive rendering module further refines the output to maintain compactness and prevent boundary overflow according to external feedback from web engine~\cite{wang2024distribution,wu2024test}.

\subsection{Perceptual Agent for Atomic Analysis}
The Perceptual Agent (PA) operates at the perceptual level to analyze the input widget image $I$. 
It follows Apple's widget design principles to decompose $I$ into atomic components, enabling subsequent MLLMs to better capture the global layout and design intent. 
The PA comprises several modules, each employing specific algorithms or external tools to perform its designated function.

\noindent \textbf{Component Extraction.} 
The input image $I$ is processed by an MLLM to detect and categorize visual components, including icons and interface primitives such as buttons, check boxes, text, and charts (e.g., bar or line). 
The complete list of component types is provided in the Appendix, defined using only the development set to avoid test data leakage. 
Each detected component element $e$ is represented as $e = [r, b, t, c]$, where $r$ denotes the cropped region, $b$ the coordinates of its bounding box, $t$ the textual description, and $c$ the component category.

\noindent \textbf{Icon Retrieval.} 
Directly prompting MLLMs to generate icons is unreliable, often causing semantic hallucinations, visual inconsistencies, and failures on small symbols with thin strokes or abstract shapes. To address this issue, we employ a retrieval-based strategy that substitutes direct generation with deterministic retrieval to preserve semantic correctness and visual fidelity. We built a collection of 50k icons in SVG format from public repositories. Each icon is first rendered into an image and then captioned by a VLM to obtain a text description, after which its visual and textual embeddings, denoted as $f_v$ and $f_t$, are extracted offline using the visual encoder $F_v$ and textual encoder $F_t$ of SigLIP~\cite{tschannen2025siglip}. Thus, each icon in the library is represented as $\{\text{SVG}^n, f_v^n, f_t^n\}_{n=1}^{N}$, where $N$ is the library size.
Given an icon element $e = [r, b, t, c]$, we compute its visual embedding $f_v^e=F_v(r)$ and textual feature $f_t^e=F_t(t)$. 
First, we conduct coarse retrieval using visual similarity by computing cosine similarity between the query visual embedding $f_v^e$ and all library visual embeddings $\{f_v^n\}_{n=1}^{N}$ to obtain the top-$K$ (K=50 in our case) candidates:
\begin{equation}
S_v(f_v^e, f_v^n) = (f_v^e)^{\top} f_v^n, \quad n = 1, \ldots, N.
\end{equation}
Next, we re-rank the coarse candidates using text-based similarity, comparing the query caption embedding $f_t^e$ with the caption embeddings $\{f_t^n\}$ of the candidate icons:
\begin{equation}
S_t(f_t^e, f_t^n) = (f_t^e)^{\top} f_t^n, \quad n \in \text{Top-}K.
\end{equation}
We retain the $\text{top-}5$ results after re-ranking.
The retrieved set $\{\text{SVG}^{i}\}_{i=1}^{5}$ represents the most visually and semantically aligned icons, which are later provided to the model for contextual exploration and integration into the generated code. Visual similarity serves as a coarse filter to ensure appearance consistency, while caption-based textual similarity re-ranks candidates to refine semantic correctness.

\begin{figure}[t]
    \centering
    \includegraphics[width=1\linewidth]{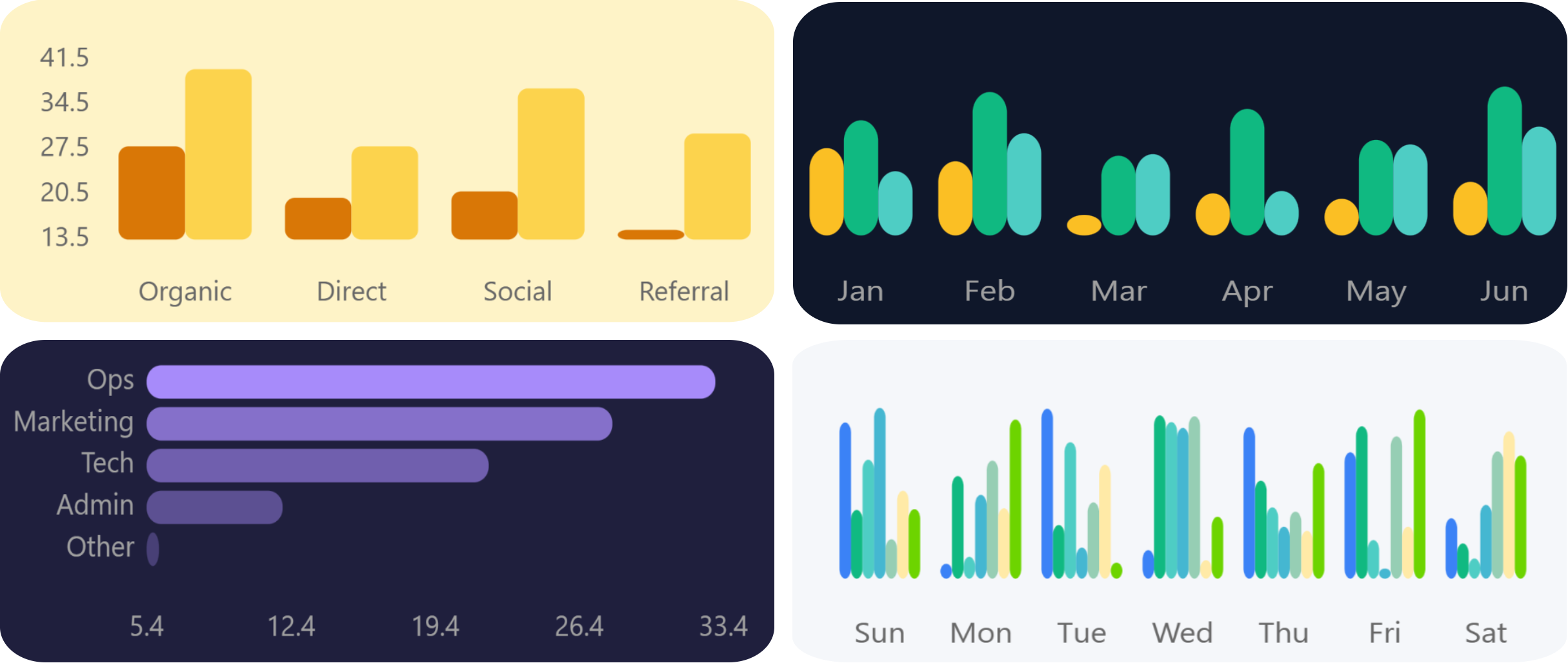}
    \caption{\textbf{Visualization of a single bar-plot template rendered diversely.} Our component templates are flexible and can be instantiated with different plot types, layouts, and color palettes.}
    \label{fig:component}
    \vspace{-0.2cm}
\end{figure}

\noindent \textbf{Reusable Component Templates.} 
For non-icon components, we define a library of reusable templates $\mathcal{T}$ written in our DSL format. 
Each template encodes the structural and functional logic of common widgets such as buttons, charts, and text blocks, \textit{exposing configurable parameters for style, data binding, and runtime behavior}. 
Given an extracted component with category $c$, cropped region $r$, and textual description $t$, the system retrieves the corresponding component template $\mathcal{T}_c$ in DSL format and prompts the MLLM with $\{r, t, \mathcal{T}_c\}$ to refine or populate the template $\mathcal{T}_c^*$ aiming to reconstruct the visual component in $r$. 
The model thus produces a customized DSL instance that preserves the predefined component structure while adapting its visual style and data semantics to the input widget. Fig.~\ref{fig:component} shows an example of instantiating a single bar-plot template into various graphs with various styles.

\noindent \textbf{Color Extraction.} 
Color is a critical element of widget design, conveying both visual hierarchy and stylistic identity. 
However, general-purpose MLLMs often fail to preserve the original color palette, resulting in noticeable deviations in tone and contrast. 
To address this limitation, we develop an image-processing module that analyzes the global color distribution of the input widget image $I$. 
The image is converted to a perceptually uniform color space after pre-processing process including filtering out transparent colors, and its pixel distribution is clustered using $K$-means to obtain the top-$K$ dominant colors and their relative proportions. 
The extracted color palette $\mathcal{P}=\{(\mu_k, w_k)\}_{k=1}^{K}$, where $\mu_k$ and $w_k$ denote the centroid and weight of each color cluster, is subsequently used to guide DSL generation and maintain stylistic consistency in the reconstructed widget. We provide more details in the Appendix.

\subsection{WidgetFactory: An End-to-End Infrastructure}

Previous UI2Code approaches primarily focus on code generation while overlooking the overall process, particularly the rendering stage. 
This limitation often results in uncontrollable and inconsistent code outputs. 
To address this issue, we introduce \textit{WidgetFactory}, a system-level infrastructure that unifies code generation, compilation, and rendering within a single end-to-end pipeline. 
It bridges the gap between perceptual understanding and executable representation, ensuring that the generated outputs are both structurally coherent and visually faithful to the input widget. 
Instead of producing lengthy and unstructured code directly from MLLMs, WidgetFactory employs a compact, interpretable, and controllable domain-specific language (WidgetDSL) specifically designed for widget representation. 
This design mitigates hallucination and redundancy while enabling fine-grained control over layout and style. 
Beyond its use in our work, WidgetFactory is designed as an easy-to-use, extensible tool to support future research and development in the Widget2Code community. We show in Appendix, WidgetFactory can be used as a data engine to synthesize data for possible development and improvement.

\begin{figure}[t]
    \centering
    \includegraphics[width=1\linewidth]{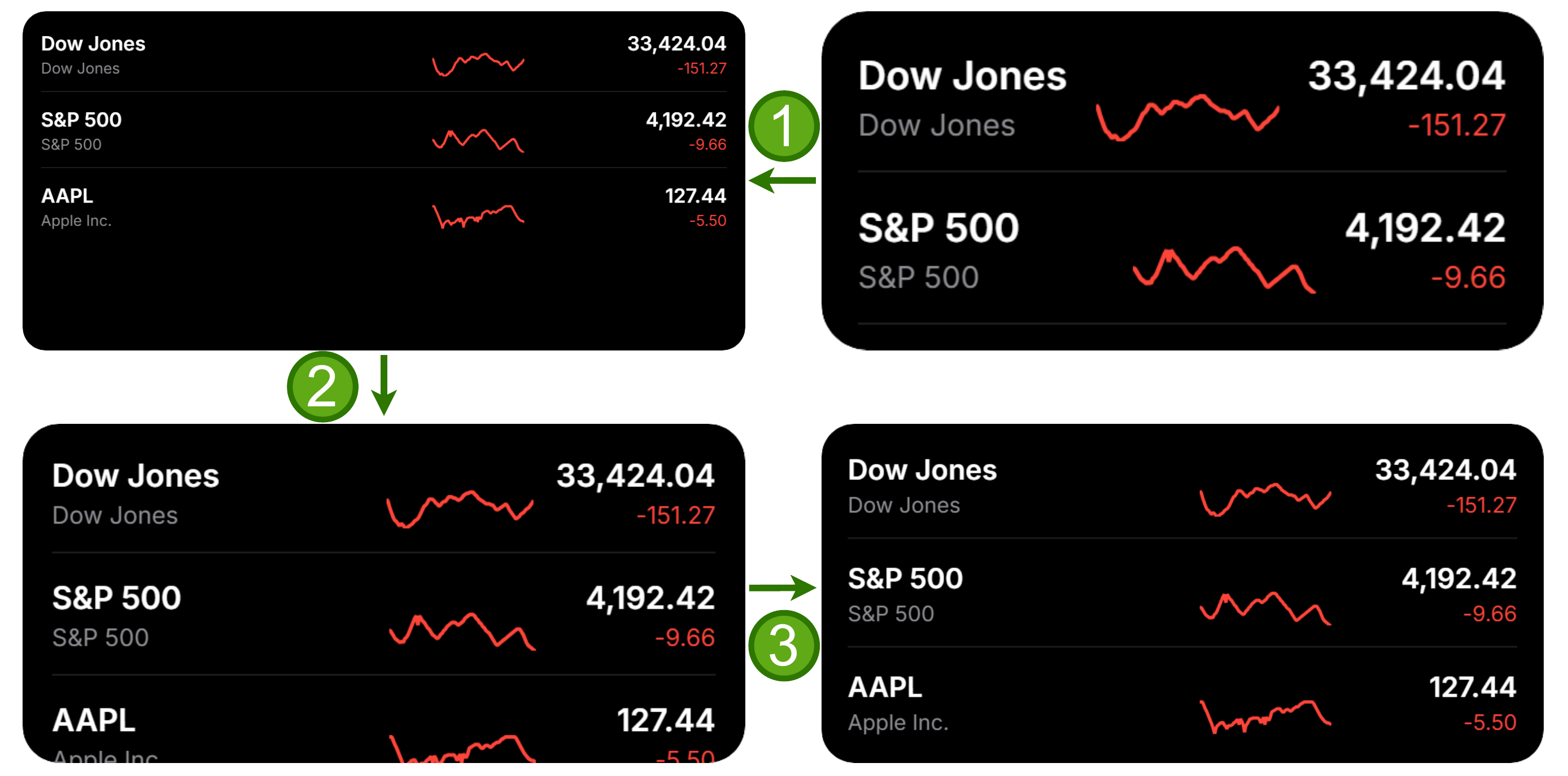}
    \vspace{-0.2cm}
    \caption{\textbf{Illustration of the adaptive rendering process.} 
1) The occluded output is first enlarged to ensure all components are visible; 
2) it is then shrunk to achieve a more compact layout, which reintroduces occlusion; 
3) finally, it is enlarged again to satisfy the spatial constraints and produce the final result.
}
    \label{fig:auto_resize}
    \vspace{-0.3cm}
\end{figure}

\noindent \textbf{WidgetDSL Design.} 
The proposed \textit{WidgetDSL}, encodes widget layouts, styles, and component hierarchies in a concise and human-readable format. 
Each widget is represented as a tree of parameterized components, where nodes correspond to functional units (e.g., icon, chart, text block) and attributes specify geometry, color, and style properties. 
The DSL syntax is intentionally minimal to facilitate generation by MLLMs while remaining expressive enough to describe diverse widget structures.

\noindent \textbf{DSL Code Generation.} The \textit{WidgetDSL} is synthesized through a multi-stage constraint-composition process that incrementally enriches the base system prompt with structured perceptual cues. Starting with a base prompt featuring the details of \textit{WidgetDSL} grammar priors, the system incrementally integrates constraints that anchor spatial, stylistic, and semantic structures in a grounded way. We inject: (1) the grounding layout constraints representing bounding boxes, textual captioning and categories from the perceptual agent ($[b, t, c]$ bounding boxes, textual descriptions and category); (2) the color palette as a ranked list of theme colors with percentages ($\mathcal{P}=\{(\mu_k, w_k)\}_{k=1}^{K}$); (3) the component specifications that regularize quantitative structure ($\mathcal{T}_c^*$), preventing hallucination in chart data and styles; (4) icon candidate sets as per-region candidate lists; (5) other dynamically inferred component types. This constraint-guided prompt chaining organizes DSL generation into a structured reasoning process, leveraging MLLMs' composition strengths while mitigating code hallucination, data invention, and layout inconsistency.

\input{sec/Algo_Fig_Tab/ablation}
\begin{figure*}[h]
    \centering
\includegraphics[width=1\linewidth]{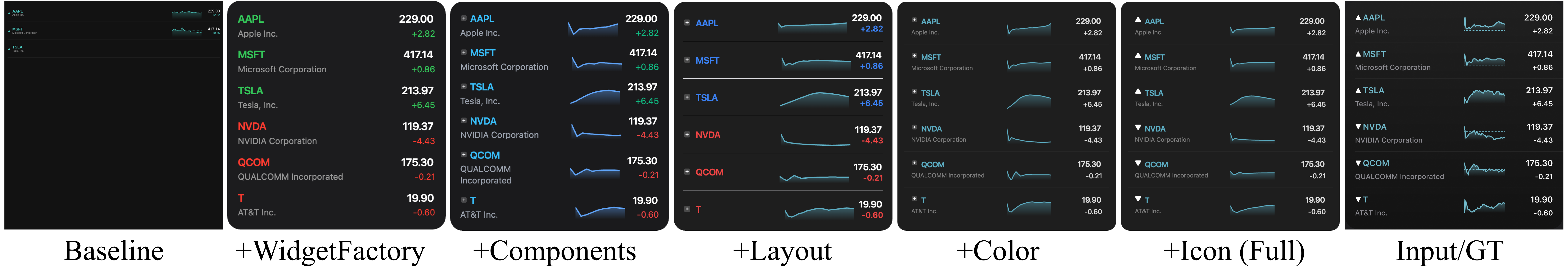}
\vspace{-0.4cm}
    \caption{\textbf{Ablation study with components adding to the system one-by-one.}
}
    \label{fig:ablation}
    \vspace{-0.3cm}
\end{figure*}
\noindent \textbf{Code Compilation.}  To bridge DSL descriptions and practical deployment, we implement a compiler system that deterministically transforms \textit{WidgetDSL} into plain web artifacts (e.g. HTML, CSS and Javascript) or framework-specific code (e.g. React). The compiler follows a two-phase DSL → AST (Abstract Syntax Tree, a hierarchical tree representation) → Target Code pipeline: (1) WidgetDSL schema validation and automatic error resolution to ensure structural validity, constructing an AST of the widget structure, and (2) recursive tree traversal to translate the abstract AST node hierarchy into its concrete code implementation. The compiler preserves hierarchical relationships and style parameters during translation, ensuring semantic equivalence between frameworks. This deterministic methodology guarantees reproducibility, and its platform-agnostic AST supports cross-platform and framework compilation.

\noindent \textbf{Adaptive Rendering.} 
To preserve aspect ratios and prevent occlusion (overflow) while maintaining spatial compactness and visual fidelity, we introduce an adaptive rendering module that searches for the closest feasible layout guided by browser feedback. 
We first calculate the aspect ratio of input widget $r = \tfrac{w}{h}$, we optimize the width $w$, with height $h = w / r$, to find a configuration where no overflow or occlusion occurs.
Formally, we seek
\begin{equation}
w^{*} = \arg\min_{w}\,[\,\Psi(w) \le 0\,], \quad h^{*} = w^{*}/r,
\end{equation}
where the violation function $\Psi(w)$ aggregates layout feedback from the rendering engine:
\begin{equation}
\Psi(w) =
\max\!\Big(
  \tfrac{C_w(w)}{V_w(w)} - 1,\;
  \tfrac{C_h(w)}{V_h(w)} - 1,\;
  \max_i \Delta_i(w)
\Big).
\end{equation}
Here, $C_w,C_h$ denote the rendered \emph{content extents}, and $V_w,V_h$ denote the \emph{viewport extents} of the container.
$\Delta_i(w)$ measures the boundary excess of each child element $i$ beyond the container padding, 
and $\Psi(w)\!\le\!0$ indicates a fully contained layout.
Starting from the natural layout size, $w$ is iteratively updated via a feedback-guided binary search: each render pass reports $\Psi(w_t)$, and $w_{t+1}$ expands or contracts until $|\Psi(w_t)| < \varepsilon$.
$\varepsilon$ is a small constant ($\varepsilon = 1/V_w$), defining the numerical tolerance for layout 
feasibility, typically corresponding to roughly one pixel relative to the container width 
to account for sub-pixel rounding errors.
This process converges to a stable configuration $(w^{*},h^{*})$ that satisfies the target aspect ratio while achieving maximal content fit and pixel-level fidelity. To match the height and width of the original widget, we apply a simple resizing step to adjust the final rendering accordingly. 
This post-processing does not degrade visual fidelity, as it only rescales the completed layout without altering its structure or style. Fig.~\ref{fig:auto_resize} shows an example of the rendering process.
As shown in Tables~\ref{tab:main_benchmark} and~\ref{tab:ablation}, our method consistently achieves a perfect geometry score, demonstrating accurate reproduction of the original widget dimensions.

%% file: sec/Algo_Fig_Tab/ablation.tex
\begin{table*}[t]
  \tabcolsep3.5pt
  \centering
    \caption{\textbf{Ablation analysis on core modules of our baseline.} Start from the baseline, and we integrate one module into the system.
  }
  \vspace{-0.3cm}
    \footnotesize
  \begin{tabular}{lccccccccccccc}
    \toprule
    \multirow{2}{*}{\textbf{Methods}}  & \multicolumn{3}{c}{\textbf{Layout}} & \multicolumn{3}{c}{\textbf{Legibility}} & \multicolumn{3}{c}{\textbf{Style}} & \multicolumn{3}{c}{\textbf{Perceptual}} & \textbf{Geometry}   \\
    \cmidrule(r){2-4} \cmidrule(r){5-7} \cmidrule(r){8-10} \cmidrule(r){11-13} \cmidrule(r){14-14}
    & \textbf{Margin} & \textbf{Content} & \textbf{Area} & \textbf{Text} & \textbf{Contrast} & \textbf{LocCon} & \textbf{Palette} & \textbf{Vibrancy} & \textbf{Polarity} & \textbf{SSIM} & \textbf{LPIPS}$\downarrow$ & \textbf{CLIP} & \\
    \midrule
    Qwen3-VL (baseline) & 64.75 & 60.15 & 69.53 & 61.17 & 60.87 & 61.12 & 47.44 & 44.50 & 54.85 & 0.703 & 0.334 & 0.800 & 95.15 \\
    \rowcolor{gray!20} + WidgetFactory & 69.97 & 64.60 & 82.46 & 67.99 & 61.53 & 57.05 & 42.36 & 42.44 & 58.81 & 0.683 & 0.339	 & 0.837 & 100 \\
    + Components & 70.83 & 64.90 & 82.30 & 67.49 & 61.44 & 57.75 & 42.61 & 41.10 & 59.30 & 0.676 & 0.344 & 0.835 &  100\\
    \rowcolor{gray!20} + Color analysis & 71.29 & 65.43 & 83.03 & 68.62 & 63.25 & 64.16 & 57.56 & 50.71 & 62.67 & 0.705 & 0.338 & 0.845  & 100\\
    + Layout & 71.49 & 65.33 & 81.92 & 68.89 & 63.64 & 64.21 & 56.10 & 49.84 & 62.54 & 0.710 & 0.340 & 0.837 & 100 \\
    \rowcolor{gray!20} + Icon (Full) & 72.15 & 66.08 & 82.24 & 70.60 & 66.20 & 64.06 & 58.09 & 51.38 & 63.28 & 0.721 & 0.335 &  0.838 & 100\\
  \bottomrule
  \end{tabular}
\label{tab:ablation}
\end{table*}

%% file: sec/Exp.tex
\subsection{Widget2Code Baseline Results and Analysis}
\noindent \textbf{Main Comparison.} 
We adopt Qwen3-VL as the base model and access it through its API throughout the entire pipeline. 
Table~\ref{tab:main_benchmark} and Fig.~\ref{fig:comparison} present comparisons against both generalized MLLMs and specialized UI2Code methods. 
Our framework effectively mitigates the identified limitations, yielding more accurate layouts, fewer missing components, and better alignment of icons, graphs, and colors. It outputs better visual fidelity and maintains aesthetic widget reconstruction.
It also preserves the input dimensions without content loss or margin overflow. 

\noindent \textbf{Modular Ablation.} 
Table~\ref{tab:ablation} summarizes the contribution of each module. 
All proposed modules lead to measurable improvements in the evaluation metrics. 
While a few combinations exhibit minor conflicts, the full integration consistently outperforms the baseline. 
Fig.~\ref{fig:ablation} provides a visual example, showing how each module contributes as expected—for instance, the \texttt{+components} module improves graph reconstruction and the \texttt{+color} module aligns the overall color palette with the input. 
Overall, these results demonstrate that our framework offers a strong and extensible baseline for future Widget2Code research.

%% file: sec/conclusion.tex
\section{Conclusion and Future Work}
In this work, we presented Widget2Code, the first systematic study of widget-focused UI2Code. 
Our work introduces an image-only widget dataset with fine-grained evaluation metrics, a comprehensive benchmarking analysis, and a modular baseline that integrates perceptual decomposition, icon retrieval, reusable visualization modules, and system-level generation through WidgetFactory. 
These contributions provide a strong foundation for reliable, interpretable, and geometry-consistent widget reconstruction. Future work may explore stronger fine-grained visual grounding for dense layouts and micro-visualizations, improved intermediate representations to reduce hallucination, and extensions to interactive or dynamic widget behaviors. 
We hope that our dataset, baseline, and system infrastructure will facilitate future progress in Widget2Code.

%% file: supp_arxiv.tex
\setlength{\abovedisplayskip}{5pt}
\setlength{\belowdisplayskip}{5pt}
\setlength{\belowcaptionskip}{-8pt}

\setcounter{section}{0}
\renewcommand{\theequation}{S\arabic{equation}}
\renewcommand{\thefigure}{S\arabic{figure}}
\renewcommand{\thetable}{S\arabic{table}}
\renewcommand{\thesection}{\Alph{section}}

\section{Summary}
In this supplementary material, we present the following additional content to complement the main paper:
\begin{itemize}
    \item Details of data curation for our widget benchmark.
    \item Details of our evaluation metrics.
    \item Details about component templates.
    \item Details about color extraction.
    \item Future directions and explorations.
    \item Illustration of prompts.
\end{itemize}

\input{supplementary/sections/sec_1}

\input{supplementary/sections/sec_2}

\input{supplementary/sections/sec_4}

\input{supplementary/sections/sec_5}
\input{supplementary/sections/sec_6}

\input{supplementary/sections/sec_3}
\input{supplementary/sections/component}


%% file: supplementary/sections/sec_1.tex
\section{Details of Data Curation}

In this section, we detail the curation of widget data that comprises the foundational dataset on which our research is based.

\subsection{Web Scraping}
To construct a diverse and high-quality dataset of widget components, we used an automated data collection pipeline to collect images of mobile widgets, targeting Dribbble, Figma and Refero. These platforms were selected due to their prevalence among professional designers, availability of the widget images data, and the high visual quality.

Data collection was performed using a Python-based crawler for Dribbble and Refero using Selenium and Beautiful Soup to parse the Document Object Model (DOM). To ensure retrieval of an adequate number and quality of assets for training generative models, the collection framework used authenticated sessions, instantiating browser instances with authenticated user credentials. Widget data from Figma were manually downloaded at high resolution, and some data was collected by researchers through manual screenshots. Post extraction, all Personally Identifiable Information was stripped from the dataset to maintain user privacy.

\subsection{Widget Extraction}
After obtaining all widget images, we must isolate the individual widget images from each screenshot. To accomplish this, we developed a multi-stage image processing pipeline, summarized in Fig.~\ref{fig:supp_data_curation} below. The pipeline is split into four distinct phases: (1) Preprocessing, (2) Union-Based Edge Detection, (3) Template Matching, (4) and Postprocessing. A visual representation is shown in the output of Fig.~\ref{fig:supp_data_curation}.

\paragraph{Preprocessing.}
We optimize the input domain, prioritizing the preservation of foreground and background separation in assets. Specifically, we have to address the challenge of heterogeneous and stylized backgrounds on which some widgets are presented upon. Since stylistic choices introduce non-uniform luminance that confounds global thresholding, we prioritize local feature extraction.

\begin{figure*}
    \centering
    \includegraphics[width=0.8\linewidth]{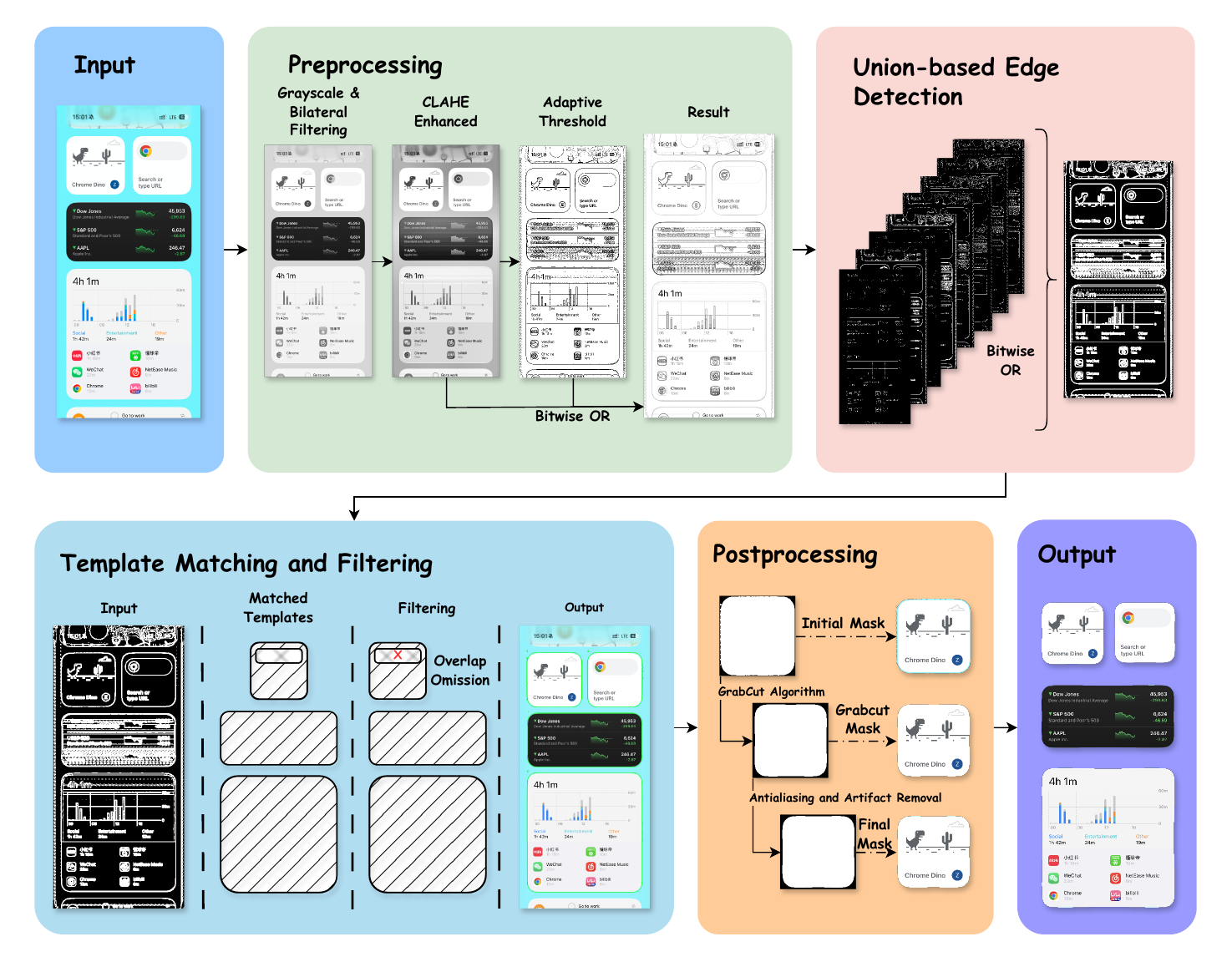}
    \caption{\textbf{The Widget Extraction Pipeline.} Visual overview of the four-stage framework transforming raw screenshots into isolated, high-fidelity assets. The process moves from Preprocessing (normalization) to Union-based Edge Detection (high-recall aggregation), followed by Template Matching (geometric verification) and Postprocessing (segmentation refinement).}
    \label{fig:supp_data_curation}
\end{figure*}

We first reduce computational complexity by converting all images to grayscale. We then conduct Bilateral Filtering ($d=5, \sigma=50)$ to preserve the sharp geometric transition of widgets while still blurring textural noise from stylized background~\cite{chi2021test}. To resolve low-contrast boundaries (e.g. dark widget on dark background), we apply Contrast Limited Adaptive Histogram (CLAHE) with a high clip limit ($12$) and localized $4 \times 4$ tiling. This ensures that edge definition is kept locally regardless of global illumination shifts. Finally, we generate a binary structural map using Adaptive Gaussian Thresholding (Block size 15, $C = 3$), which we found to best isolate widget boundaries based on neighborhood intensity differences. 

Since some edge detectors in the Union-Based Edge Detection stage requires textural nuance for derivative-based edge detection (e.g. Sobel), we can't use the binary structural map from the Adaptive Gaussian Thresholding in isolation (it discards the textural information). Therefore, we fuse the original CLAHE signal with the result of the Adaptive Thresholding using a Bitwise OR operation to create a hybrid representation. By retaining both binary location data and textural nuance, we ensure the input is robust enough to satisfy the diverse requirements of the subsequent ensemble edge detection stage.

\paragraph{Union-based Edge Detection}
Given the high variance in widget presentation, no single edge detection algorithm was found to yield sufficient and accurate recall across the entire dataset. While standard intensity-gradient methods capture sharp boundaries, they fail on isoluminant regions (differing hue, identical brightness); conversely, chromatic and morphological detectors resolve color and shape-based edges but often lack fine-grained precision. To maximize coverage, we aggregate nine parallel detectors categorized by their dominant modality:

\begin{itemize}
\item \textbf{Multi-Scale Intensity Analysis.} We run three parallel Canny instances using progressively larger Gaussian blur kernels ($3\times3$, $5\times5$, $7\times7$) to capture edges at varying spatial frequencies. We utilize aggressively minimized hysteresis thresholds (Lower: $1-5$ / Upper: $15-30$) compared to standard usage ($100/200$) to detect subtle 1-pixel borders, accepting increased noise for higher recall.

\item \textbf{Derivative \& Structural Analysis.} To capture non-step edges, we employ a Laplacian operator ($k=3$, threshold=5) to identify zero-crossings and a 64-bit floating-point Sobel operator to capture directional gradients independent of magnitude. Additionally, a Morphological Gradient ($3\times3$ rectangular kernel, threshold=10) is computed to highlight boundaries based on local structural range rather than directional intensity, providing resilience against lighting artifacts.

\item \textbf{Chromatic Analysis.} To address the iso-luminant problem, we apply Canny detection strictly to the Saturation ($20/80$ threshold) and Value ($15/60$ threshold) channels of the HSV color space. This successfully isolates boundaries defined by shifts in color purity that vanish in grayscale conversion.

\end{itemize}

The binary outputs of these nine detectors are fused via another Bitwise OR. This union-based approach prioritizes recall, ensuring that any boundary detected by at least one modality is preserved. Finally, to ensure topological consistency for contour extraction, the composite map undergoes Morphological Closing ($3\times3$ kernel) to bridge fragmented segments, followed by minimal Dilation ($3\times3$) to reinforce edge connectivity.

\paragraph{Template Matching \& Geometric Filtering}
While high in recall, the Union-based Edge Detection also contains noise and non-semantic contours (e.g. text blocks, icons). To extract valid widgets, we undergo a geometric matching stages assuming widgets are shaped as rounded rectangles; circular widgets are not considered. We synthesized a multi-scale template library spanning dimensions from $80$px to $2000$px, utilizing \textit{non-linear quantization} (step sizes ranging from $15$px for icons to $125$px for containers) to use for template matching.

To distill semantic components~\cite{ye2024bayes}, we validate contours against a synthesized library of rounded rectangles using \textit{Hu Moments}. This metric's scale-invariance decouples topology from resolution, enabling a single geometric primitive to verify diverse widgets ($80$--$2000$px) regardless of pixel dimensions. We enforce adaptive similarity thresholds ($I_1 < 0.02$ standard; $< 0.05$ large) to accommodate internal complexity, prune artifacts via geometric constraints (Solidity $>0.5$), and eliminate duplicates via greedy non-maximum suppression ($30\%$ overlap).

\paragraph{Postprocessing \& Refinement.}
To refine coarse geometric matches to pixel-perfect boundaries, we initialize the \textbf{GrabCut} algorithm, seeding the algorithm with the template-matched mask to provide a strong spatial prior. We restrict the process to two iterations, utilizing Gaussian Mixture Models to tightly adapt contours to local color distributions while preventing potential bleeding into the background. Following segmentation, we apply two distinct refinement passes. First, for \textbf{Anti-Aliasing}, we synthesize a continuous alpha matte using a Euclidean Distance Transform, applying stratified blending ($40\%/20\%$ weights) within a $2$-pixel edge zone to smooth jagged boundaries. Second, for Artifact Removal, we excise background ``halos'' using a color-difference heuristic; edge pixels are compared to the adjacent background, and those with a Euclidean distance $<30$ are removed to ensure the final transparency is free of color bleeding.

The cumulative result of this extraction framework is a standardized dataset of high-fidelity, transparent RGBA widget assets, each cropped to its minimal bounding box. These isolated components serve as the foundational visual dataset for our subsequent research.

%% file: supplementary/sections/sec_2.tex
\section{Details of Evaluation Metrics}

In this section, we provide detailed derivations of the proposed evaluation metrics, following Apple’s official widget design principles. Note that in the main paper, most metrics are reported as similarity scores between the prediction and the ground truth. \textbf{For clarity, we first derive each metric in its difference form and then convert it into a similarity measure (e.g., difference → similarity, asymmetry → symmetry).}

\noindent To convert each difference-based layout metric into a bounded similarity score, we apply an exponential transformation of the form
\begin{equation}
\text{Sim}(v) = \exp(-\, v / s),
\end{equation}
where $v$ is the raw difference value and $s$ is a scale parameter controlling 
the decay rate. 
This mapping yields a similarity score in $[0,1]$ that is 
maximal for perfect matches and decreases smoothly as the deviation increases.
Then, we multiply all the scores by 100, making them fall in the range of $[0,100]$

\subsection{Layout}
\paragraph{Structural mask and margin box.}
To evaluate layout fidelity, we operate on a geometry-only representation of the widget that is invariant to color, texture, and style. 
We therefore extract a binary \emph{structural mask} by applying an edge detector to the input image and dilating the resulting edges. 
This produces a stable outline of the widget's visible structure while suppressing interior appearance details. 
All layout metrics are subsequently computed on this structural mask.

\begin{figure}
    \centering
    \begin{subfigure}{\linewidth}
        \centering
        \includegraphics[width=1\linewidth]{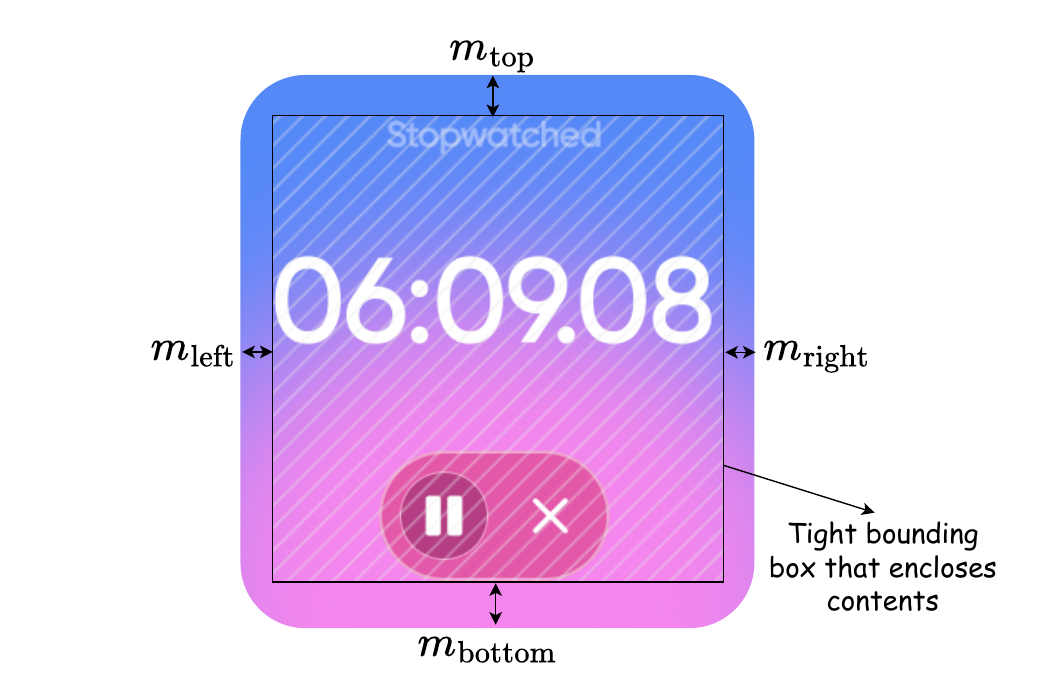}
    \end{subfigure}
    \caption{\textbf{An example to show the margins of the widget.}}
    \label{fig:supp_margin}
\end{figure}

Given a structural mask $M \in \{0,1\}^{H \times W}$ with support 
\begin{equation}
    \mathcal{S}(M)=\{(y,x)\mid M(y,x)=1\},
    \label{eq: supp_diff_to_sim}
\end{equation}

the four margins are defined directly from the extremal coordinates of the mask:
\begin{align}
m_{\text{top}}    &= \min_{(y,x)\in\mathcal{S}(M)} y, \\
m_{\text{bottom}} &= H - 1 - \max_{(y,x)\in\mathcal{S}(M)} y, \\
m_{\text{left}}   &= \min_{(y,x)\in\mathcal{S}(M)} x, \\
m_{\text{right}}  &= W - 1 - \max_{(y,x)\in\mathcal{S}(M)} x.
\end{align}
We denote the margin vector as
\begin{equation}
m(M) = (m_{\text{top}},\, m_{\text{bottom}},\, m_{\text{left}},\, m_{\text{right}}),
\end{equation}
which serves as the basis for all margin-based layout metrics.
Intuitively, we compute the tightest bounding box that encloses all structural (non-background) pixels in the widget, and $m(M)$ records its distances to the four image boundaries. Fig.~\ref{fig:supp_margin} shows an example of the computed margin distances.

\paragraph{Margin symmetry (Margin).}
Balanced padding is a core widget design principle, and uneven deviations across the four margins indicate misalignment in the reconstructed layout. 
To quantify this imbalance, we compare the ground-truth and generated margin vectors. 
Let $m(M)\in\mathbb{R}^4$ denote the margin vector and define
\begin{equation}
\Delta m = \lvert\, m(M_{gt}) - m(M_{gen}) \,\rvert.
\end{equation}
Let $\mu$ and $\sigma$ be the mean and standard deviation of the four entries of $\Delta m$. 
With a small numerical threshold $\varepsilon = 10^{-6}$, the margin asymmetry metric is
\begin{equation}
\text{MarginAsym} =
\begin{cases}
0, & \mu < \varepsilon,\\[3pt]
\sigma / \mu, & \text{otherwise},
\end{cases}
\end{equation}
which increases as the margin deviations become uneven across sides. The asymmetry is then converted into margin symmetry using Eq.~\ref{eq: supp_diff_to_sim}.

\paragraph{Content Aspect Ratio Similarity (Content).}
Widgets are designed to scale across size classes while preserving the proportional structure of their content. 
To measure whether the reconstructed widget maintains this proportionality, we compare the aspect ratios of the tight bounding boxes enclosing the structural masks. 
For a mask $M$, let
\begin{equation}
h(M) = y_{\max} - y_{\min} + 1, \qquad
w(M) = x_{\max} - x_{\min} + 1,
\end{equation}
where the extremal coordinates are computed over $\mathcal{S}(M)=\{(y,x)\mid M(y,x)>0\}$. 
The content aspect ratio is
\begin{equation}
\text{AR}(M) = \frac{w(M)}{h(M)}.
\end{equation}
Given ground-truth and generated masks, the \emph{content aspect ratio difference} is
\begin{equation}
\text{ContentAspectDiff}
= \bigl|\, \log(\text{AR}(M_{gt}) / \text{AR}(M_{gen})) \,\bigr|,
\end{equation}
which is zero when the proportional shape is preserved and increases as the reconstruction becomes stretched or compressed. The ratio difference is then converted into content aspect ratio similarity using Eq.~\ref{eq: supp_diff_to_sim}.

\paragraph{Area Ratio Similarity (Area).}
Balanced visual hierarchy is a key widget design principle: internal elements should preserve their relative visual weight when the layout is reconstructed. 
We measure this by comparing the normalized area distribution of connected components in the structural masks. 
Let $\{A_i^{gt}\}$ and $\{A_j^{gen}\}$ denote the areas of the connected components (after discarding tiny regions). 
We compute the normalized mean area for each mask,
\begin{equation}
r_{gt} = \frac{\frac{1}{n_{gt}}\sum_i A_i^{gt}}{\sum_i A_i^{gt}},
\qquad
r_{gen} = \frac{\frac{1}{n_{gen}}\sum_j A_j^{gen}}{\sum_j A_j^{gen}},
\end{equation}
and define the \emph{area ratio difference} as
\begin{equation}
\text{AreaRatioDiff} = \bigl|\, r_{gt} - r_{gen} \,\bigr|.
\end{equation}
This value increases when the reconstruction disproportionately enlarges or shrinks certain components, altering the intended visual balance of the widget. The difference is then converted into area ratio similarity using Eq.~\ref{eq: supp_diff_to_sim}.

Both \emph{Margin Symmetry} and \emph{Content Aspect Ratio Similarity} operate at the global level: 
they treat the entire structural region as a single entity and measure how its overall position and shape differ between the ground-truth and generated widgets, using only the outer margin bounding box. 
In contrast, \emph{Area Ratio Similarity} examines the internal structure within this box by comparing the normalized area distribution of connected components, thereby capturing changes in visual hierarchy that global metrics cannot detect.

\subsection{Legibility}

Widgets must remain quickly readable at a glance, and Apple’s guideline emphasizes clear typography, adequate contrast, and preservation of textual meaning. Our legibility metrics evaluate whether a reconstructed widget maintains (1) correct text content, (2) comparable global contrast, and (3) sufficient local text contrast for readability.

\paragraph{Text Jaccard (Text).}
To assess the semantic consistency of textual content, we apply OCR (EasyOCR) to both 
the ground-truth and generated widgets and compare their extracted word sets.
Let $W_{gt}$ and $W_{gen}$ denote the sets of OCR-detected words.
The text similarity is given by the Jaccard index
\begin{equation}
\text{TextJaccard} = 
\frac{|\, W_{gt} \cap W_{gen} \,|}{|\, W_{gt} \cup W_{gen} \,|}.
\end{equation}
A higher value indicates better preservation of the widget's textual meaning.

\paragraph{Contrast similarity (Contrast).}
To evaluate global readability, we compare the overall luminance contrast 
between the two widgets. 
Let $L$ be the grayscale image, and let $L_{5}$ and $L_{95}$ denote its 
5th and 95th percentile luminance values. 
The contrast ratio is
\begin{equation}
C(L) = \frac{L_{95} + 0.05}{L_{5} + 0.05}.
\end{equation}
The global contrast deviation is then
\begin{equation}
\text{ContrastDiff} = \bigl|\, C(L_{gt}) - C(L_{gen}) \,\bigr|.
\end{equation}
The difference is then converted into similarity using Eq.~\ref{eq: supp_diff_to_sim}.

\paragraph{Local Contrast Similarity (LocCon).}
To measure readability within text regions, we compute contrast ratios 
inside the OCR-detected bounding boxes. 
Let $C_{\text{local}}(L)$ denote the mean contrast over all detected text regions.
The local contrast deviation is
\begin{equation}
\text{ContrastLocalDiff} =
\bigl|\, C_{\text{local}}(L_{gt}) - C_{\text{local}}(L_{gen}) \,\bigr|.
\end{equation}
If text is detected in only one of the two widgets, a fixed penalty is applied.
The difference is then converted into similarity using Eq.~\ref{eq: supp_diff_to_sim}.

\subsection{Style}

Widgets must maintain consistent visual styling—particularly color theme, saturation levels, and foreground–background contrast—to preserve brand identity and aesthetic coherence.
The following metrics measure style fidelity in terms of global hue statistics, saturation distribution, and luminance polarity.

\paragraph{Palette Distance.}
We compare the global color themes of the two widgets by computing the
Earth Mover's Distance between their hue histograms.
Let $h_{gt}$ and $h_{gen}$ denote the normalized histograms over $B$ bins 
in HSV hue space.
The palette deviation is
\begin{equation}
\text{EMD}_{\text{hue}} = W_1(h_{gt}, h_{gen}),
\end{equation}
where $W_1$ is the 1D Wasserstein distance.
The palette similarity is then
\begin{equation}
\text{PaletteDistance} = \exp\!\left(-\,\frac{\text{EMD}_{\text{hue}}}{\alpha}\right),
\end{equation}
with scale $\alpha$ controlling sensitivity.

\paragraph{Vibrancy Consistency.}
Let $s_{gt}$ and $s_{gen}$ denote the normalized histograms of saturation 
values in HSV space.
The vibrancy deviation is
\begin{equation}
\text{EMD}_{\text{sat}} = W_1(s_{gt}, s_{gen}),
\end{equation}
and the corresponding similarity is
\begin{equation}
\text{Vibrancy} = \exp\!\left(-\,\frac{\text{EMD}_{\text{sat}}}{\beta}\right).
\end{equation}
Higher values indicate better alignment in saturation distribution and 
overall vividness.

\paragraph{Polarity Consistency.}
Let $L$ be the grayscale image, and let 
$\text{bg}(L)$ be the median luminance (approximate background) and 
$\text{fg}(L)$ the mean luminance of the darkest $10\%$ pixels 
(approximate foreground). 
The polarity sign is
\begin{equation}
p(L) = \operatorname{sign}\!\bigl(\text{bg}(L) - \text{fg}(L)\bigr).
\end{equation}
Polarity consistency is defined as
\begin{equation}
\text{PolarityConsistency} = 
\begin{cases}
\exp(-\,\gamma \, |\Delta|), & p(L_{gt}) = p(L_{gen}),\\
0, & \text{otherwise},
\end{cases}
\end{equation}
where $\Delta = \bigl[(\text{bg}-\text{fg})_{gt} - (\text{bg}-\text{fg})_{gen}\bigr]$ 
captures magnitude differences and $\gamma$ is a scale factor.

\subsection{Human evaluation}
To validate that our designed evaluation metrics align with human perception, we conduct user studies. Specifically, we recruit 25 participants to rank the results among outputs across models (100 test widgets in total). We include non-hallucination.
The resulting ranking in Table~\ref{tab:human} matches Table~1 (Ours$>$Gemini$>$Qwen3$>$UIUC). We further compute the ranking correlation between humans and our metric for each widget and average, which is consistently high.

\begin{table}[h]
\centering
\small
    \tabcolsep3pt
    \begin{tabular}{c|cccc}
        \toprule
       Human Rank (1:best, 4:worst) & Lay. & Legi. & Sty. & No-Hallu. \\
        \midrule
        Qwen3-VL & 2.48 & 2.60& 2.49 & 2.43 \\
        \rowcolor{gray!20} Gemini & 2.40 & 2.35& 2.32 & 2.39 \\
        UIUG & 3.71 & 3.64 & 3.74 & 3.75 \\
        \rowcolor{gray!20} Qwen3-VL + Ours & 1.42 & 1.41 & 1.45 & 1.43 \\
        \midrule
        Metric--Human correlation & 0.93 & 0.89 & 0.94 & -\\
        \bottomrule
    \end{tabular}
    \caption{Alignment with human evaluation.}
    \label{tab:human}
\end{table}

%% file: supplementary/sections/sec_4.tex
\section{Details about Component Templates}

The component templates we summarized from the development set include the following items:

\begin{nolinenumbers}
\begin{multicols}{3}
\begin{itemize}
    \item Bar chart
    \item Line chart
    \item Pie chart
    \item Radar chart
    \item Button
    \item Checkbox
    \item Divider
    \item Image
    \item Indicator
    \item Progress bar
    \item Progress ring
    \item Slider
    \item Switch
    \item Text
\end{itemize}
\end{multicols}
\end{nolinenumbers}

\begin{figure}
    \centering
    \begin{subfigure}{\linewidth}
        \centering
        \includegraphics[width=1\linewidth]{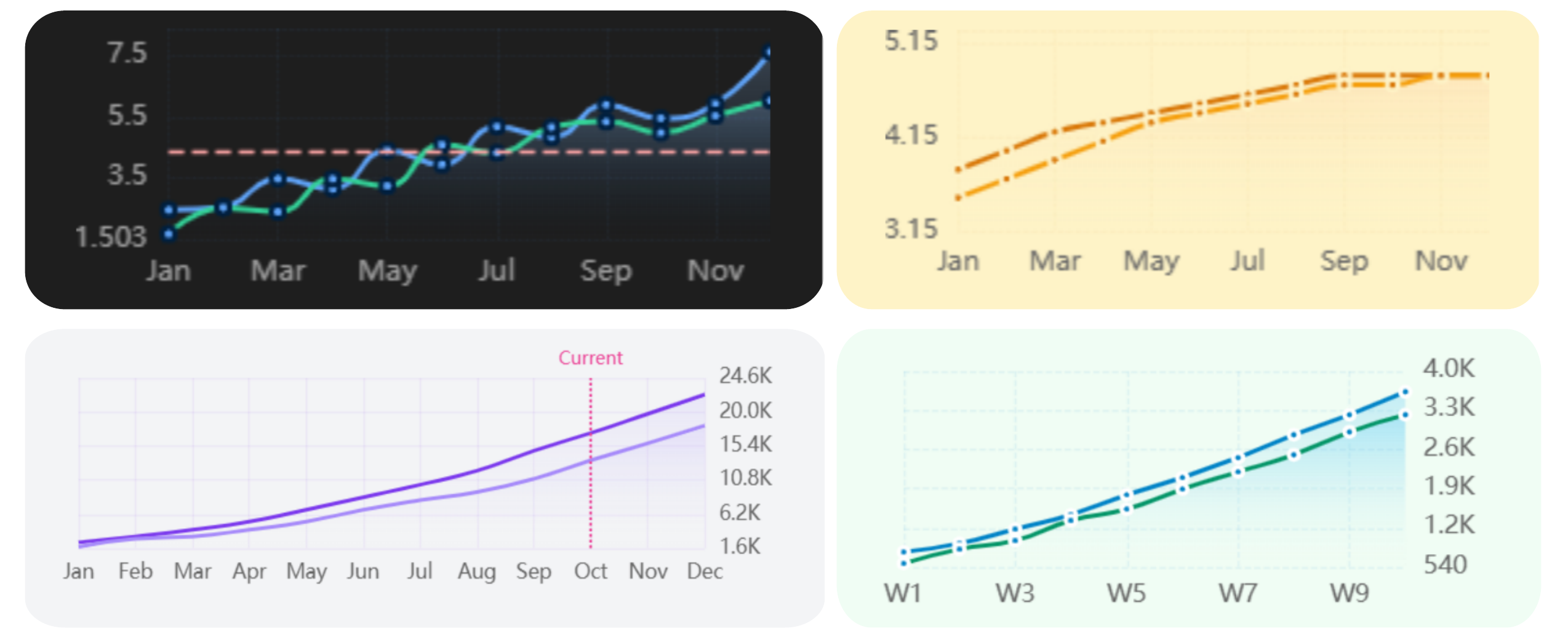}
        \caption{Line chart}
        \vspace{0.4cm}
    \end{subfigure}
    \begin{subfigure}{\linewidth}
        \centering
        \includegraphics[width=1\linewidth]{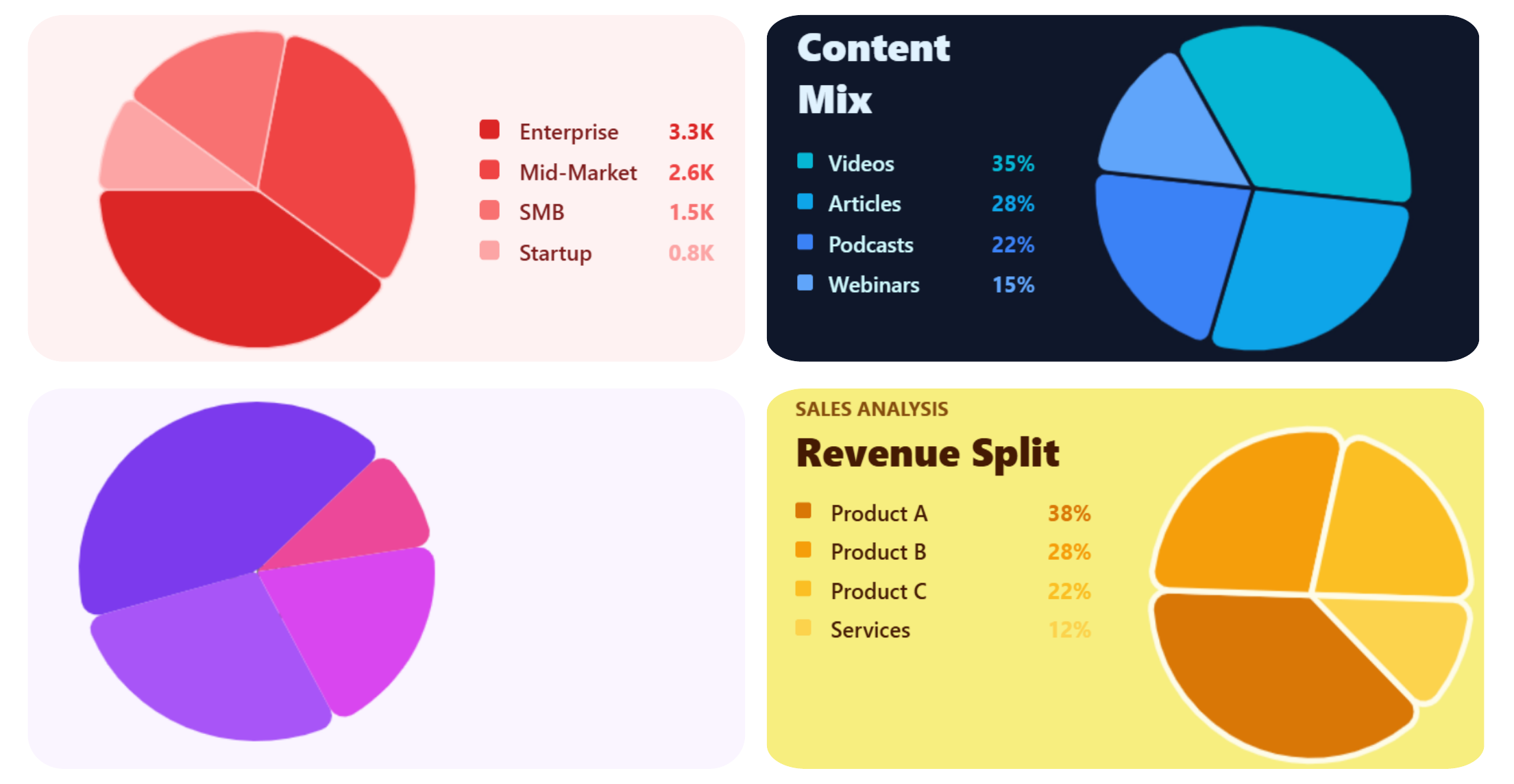}
        \caption{Pie chart}
        \vspace{0.4cm}
    \end{subfigure}
    \begin{subfigure}{\linewidth}
        \centering
        \includegraphics[width=1\linewidth]{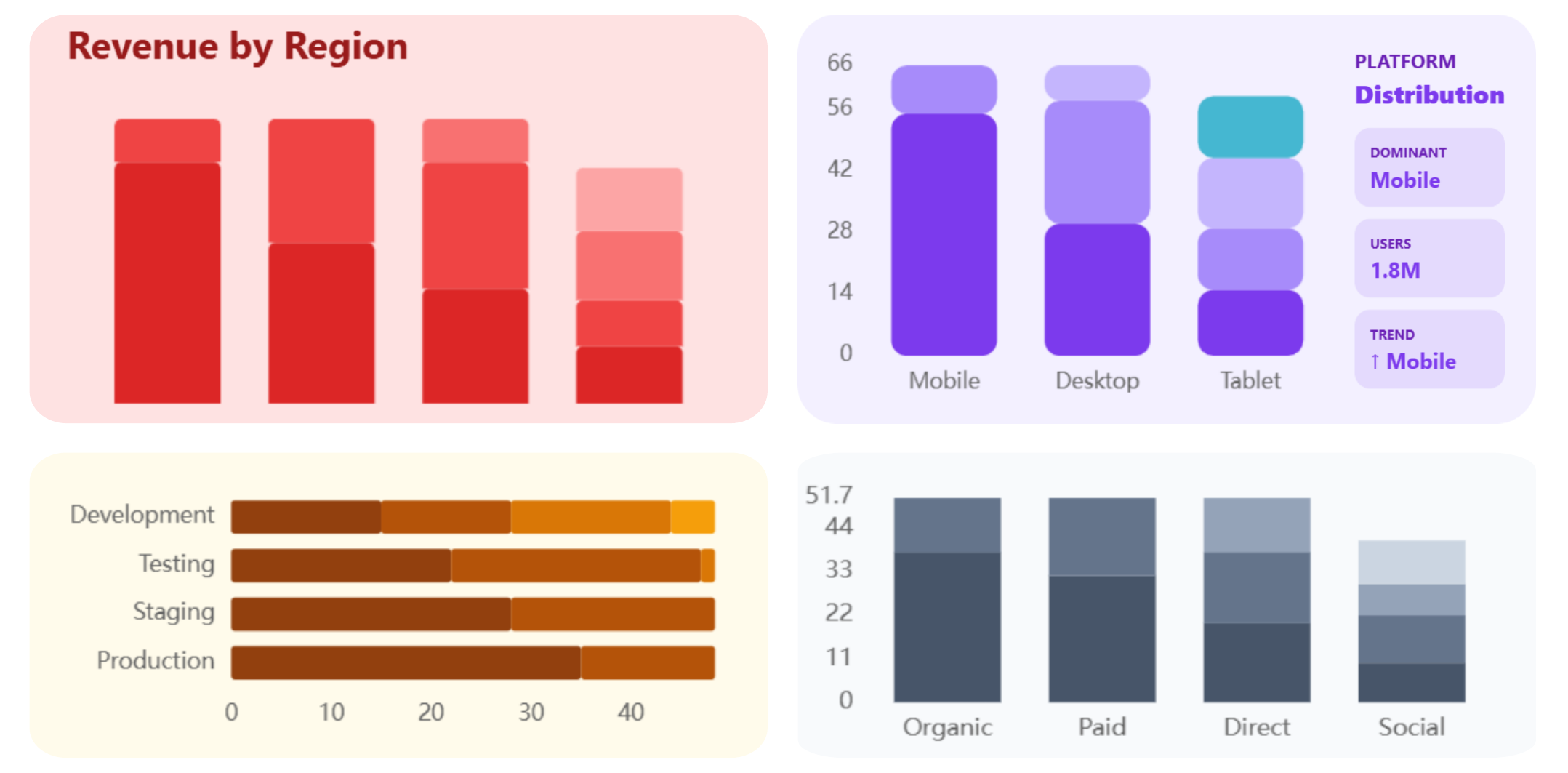}
        \caption{Stacked bar chart}
    \end{subfigure}
    \caption{\textbf{Additional rendered examples from single-component templates.} We present more visual samples derived from the line chart, pie chart, and stacked bar chart templates, illustrating broad coverage across multiple graph types.}
    \label{fig:supp_component}
\end{figure}

\noindent While the main paper showcased examples rendered from one bar-plot template, Fig.~\ref{fig:supp_component} illustrates additional results derived from multiple template types. Collectively, these examples demonstrate that our component templates are flexible and visually robust across diverse data streams and design aesthetics.

%% file: supplementary/sections/sec_5.tex
\section{Details of Color Extraction}
The algorithm extracts dominant colors from a widget image by performing k-means color quantization followed by a full-resolution reassignment step. The method operates by (1) loading and preprocessing the image, (2) clustering a sampled subset of pixels to identify representative color centers, and (3) assigning all original pixels to their nearest cluster to accurately compute global color proportions. The final output consists of the top colors expressed in hexadecimal form along with their relative frequencies.
\paragraph{\textbf{Key Components.}} The procedure can be decomposed into the following three main stages:
\begin{itemize}
\item \textbf{Image Preprocessing and Sampling.} The input image is loaded and converted to RGB format, with transparent pixels optionally removed when an alpha channel exists. To improve computational efficiency on high-resolution images, up to a pre-set fixed number of pixels are randomly sampled while preserving the statistical distribution of the colors. For computational efficiency, we optionally constrain the total number of processed pixels to \(1{,}000{,}000\) (\(1000 \times 1000\)), matching the maximum expected resolution of a typical widget image. This cap accelerates clustering on high-resolution assets but can be relaxed when full-resolution fidelity is required.
\item \textbf{K-means Color Quantization.} After transparent colors are filtered, the remaining pixel distribution is clustered using k-means to obtain the top dominant colors and their relative proportions. The sampled pixels are reshaped into a matrix of dimension \(M \times 3\), where \(M\) is the number of sampled pixels, 3 represents the Red, Green, and Blue (RGB) color channels for each pixel. 
\item \textbf{Full-pixel Reassignment and Color Palette.} To ensure accurate color frequency estimation, every pixel in the original image is reassigned to its nearest cluster center based on squared Euclidean distance. Cluster membership counts are aggregated across all pixels and sorted in descending order. The resulting statistics correspond to the color palette $\mathcal{P}=\{(\mu_k, w_k)\}_{k=1}^{K}$, where $\mu_k$ and $w_k$ denote the centroid and weight of each color cluster. From this palette, the top $n$ \ colors (In our implementation, $n$ \ is set to $8$\, which provides a compact yet expressive summary of the widget’s color distribution without overwhelming the downstream representation) and their percentage contributions—expressed in the hexadecimal RGB format—are obtained. This representation captures the dominant chromatic structure and is subsequently used to guide DSL generation while maintaining stylistic consistency.
\end{itemize}

\begin{figure}
    \centering
    \begin{subfigure}{\linewidth}
        \centering
        \includegraphics[width=1\linewidth]{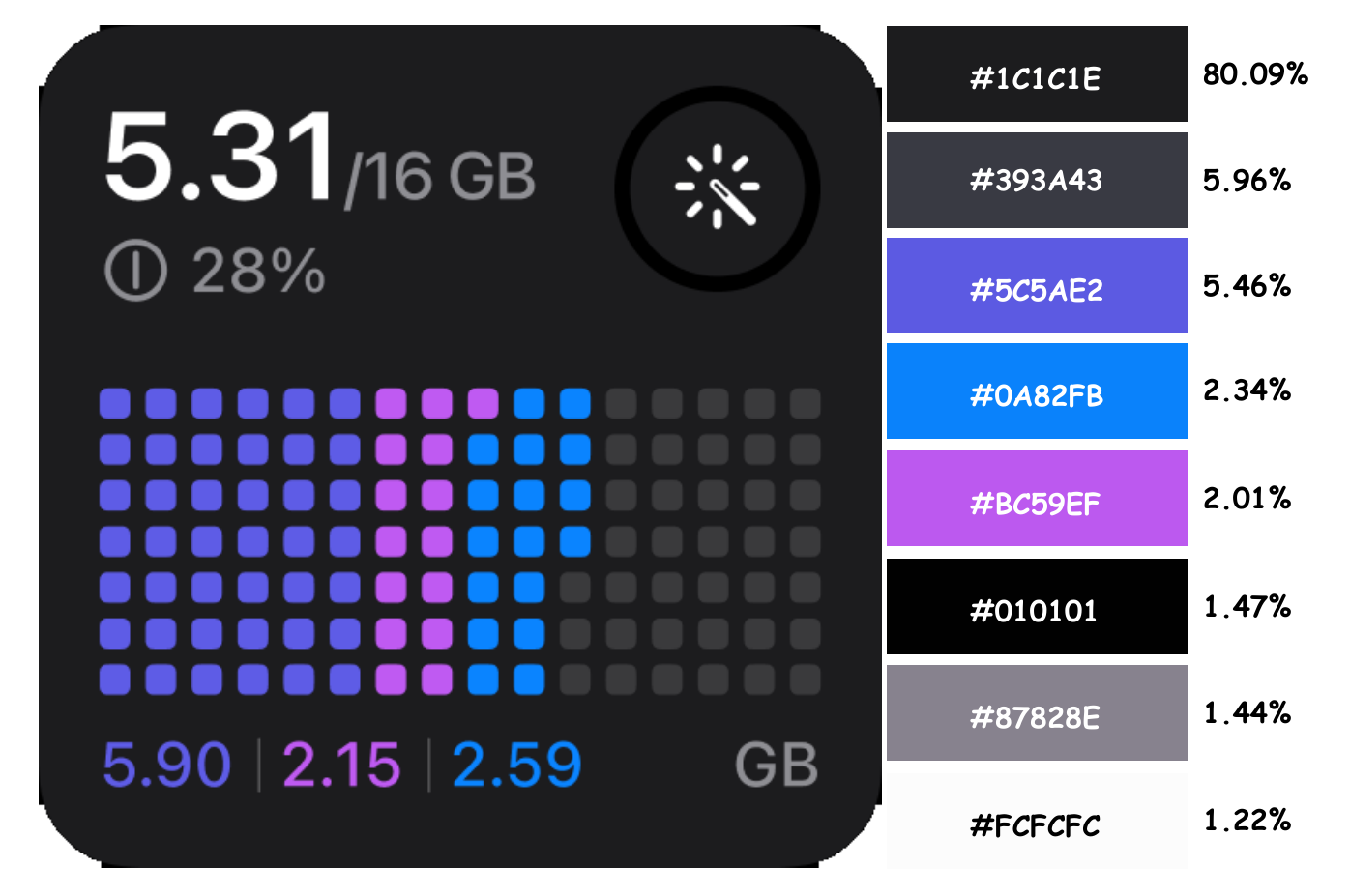}
    \end{subfigure}
    \caption{\textbf{An example of color extraction.}}
    \label{fig:supp_color}
\end{figure}

Directly counting pixel colors in a UI image is ineffective because modern interfaces contain anti-aliasing, gradients, shadows, and compression noise that produce an extremely large number of distinct RGB values. These raw pixel variations do not correspond to meaningful stylistic choices and therefore cannot be interpreted as a stable color palette. K-means clustering provides a principled, unsupervised method for reducing this high-variance color space into a small set of representative chromatic centroids. By grouping perceptually similar pixels, K-means recovers the underlying dominant colors while suppressing noise arising from rendering artifacts. This yields a compact and robust estimate of the UI’s true palette distribution, which can be reliably conveyed to downstream modules for style-consistent code generation.
Fig.~\ref{fig:supp_color} shows an example of color extraction output. 

%% file: supplementary/sections/sec_6.tex
\section{Future directions and explorations.}

\begin{figure*}[t]
  \centering
  \includegraphics[width=0.75\textwidth]{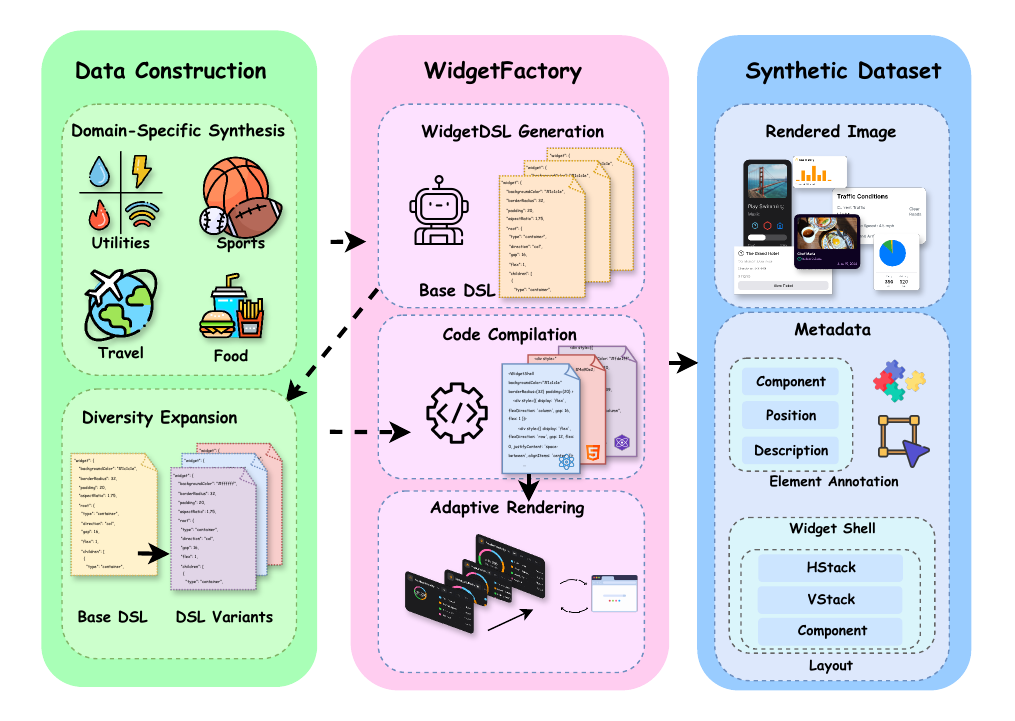}
  \caption{\textbf{The WidgetFactory Synthetic Generation Pipeline.} 
  The framework utilizes domain-conditioned synthesis and diversity expansion to generate WidgetDSL specifications. 
  These are compiled and rendered to produce a multimodal dataset comprising high-fidelity images, element-level annotations, and hierarchical layout data.}
  \label{fig:data_engine}
\end{figure*}

In addition to its core functionality as a generation and rendering infrastructure, \textbf{WidgetFactory} can also serve as a versatile \textbf{data engine}. By composing diverse widget specifications using WidgetDSL, our system enables the generation of large-scale synthetic datasets with controllable variation in layout, style, hierarchy, and component types. Crucially, each synthetic instance is synthesized as an aligned quadruplet $(I, L, D, C)$—comprising the visual Image, spatial Layout, structural DSL, and executable Code, facilitating supervised learning, pretraining, or benchmarking in Widget2Code-related tasks. This capability supports future research on UI grounding, layout reasoning, icon retrieval, and robust code generation, and contributes to the development of scalable and reproducible pipelines for the Widget2Code community. In the following, we give examples of potential tasks and the data generation process. \textbf{Please note that it is a preliminary attempt at using WidgetFactory as a data engine for possible improvement.}

\subsection{Synthetic Data Generation via WidgetFactory}

\paragraph{Synthesis of WidgetDSL Specifications}
The initialization stage employs a high-throughput, LLM-driven synthesis engine to programmatically populate the dataset. Iterating through target domains (e.g., Utilities, Social, Retail), the system first leverages Large Language Models to generate a massive corpus of diverse natural language widget descriptions. These high-level textual concepts are subsequently transpiled into formal WidgetDSL specifications. Crucially, this translation is conditioned on reference widgets—existing structural exemplars that serve as few-shot prompts—to ground the generation in valid syntax and realistic hierarchies. The result is a domain-partitioned library of executable DSLs that preserves semantic diversity while guaranteeing structural renderability.



\paragraph{Controlled Mutation for Diversity Expansion}
To ensure data variety, we expand the synthesis corpus by applying controlled, rule-driven mutations to both the render-ready DSLs. A mutator generates diversity through deterministic theme-based transformations. Each base DSL is expanded into multiple stylistic variants---including light, dark, colorful, glassmorphism, and minimal themes---using a mutation palette that specifies allowable changes to layout, typography, colors, and chart attributes. Batch outputs are validated and yield a large corpus of unique widget specifications.

\paragraph{WidgetFactory and Synthetic Dataset}
The WidgetFactory compiles each WidgetDSL into executable JSX and renders it into pixel-accurate artifacts using an instrumented headless browser pipeline. The renderer then loads each DSL, generates JSX and layout components, and executes them in a headless Browser instance. During rendering, the system captures high-fidelity PNG screenshots and computes DOM-aligned bounding boxes for all visible elements using the same structural identifiers introduced during synthesis and preserved throughout compilation. Parallel execution is controlled via a concurrency parameter, balancing throughput against resource usage. Each synthetic widget’s output directory contains the full artifact bundle---the final rendered image, layout and element-annotation metadata---providing deterministic traceability from specification to rendered image. Fig.~\ref{fig:data_engine} illustrates this process.

\begin{figure}[t]
  \centering
  \includegraphics[width=\linewidth]{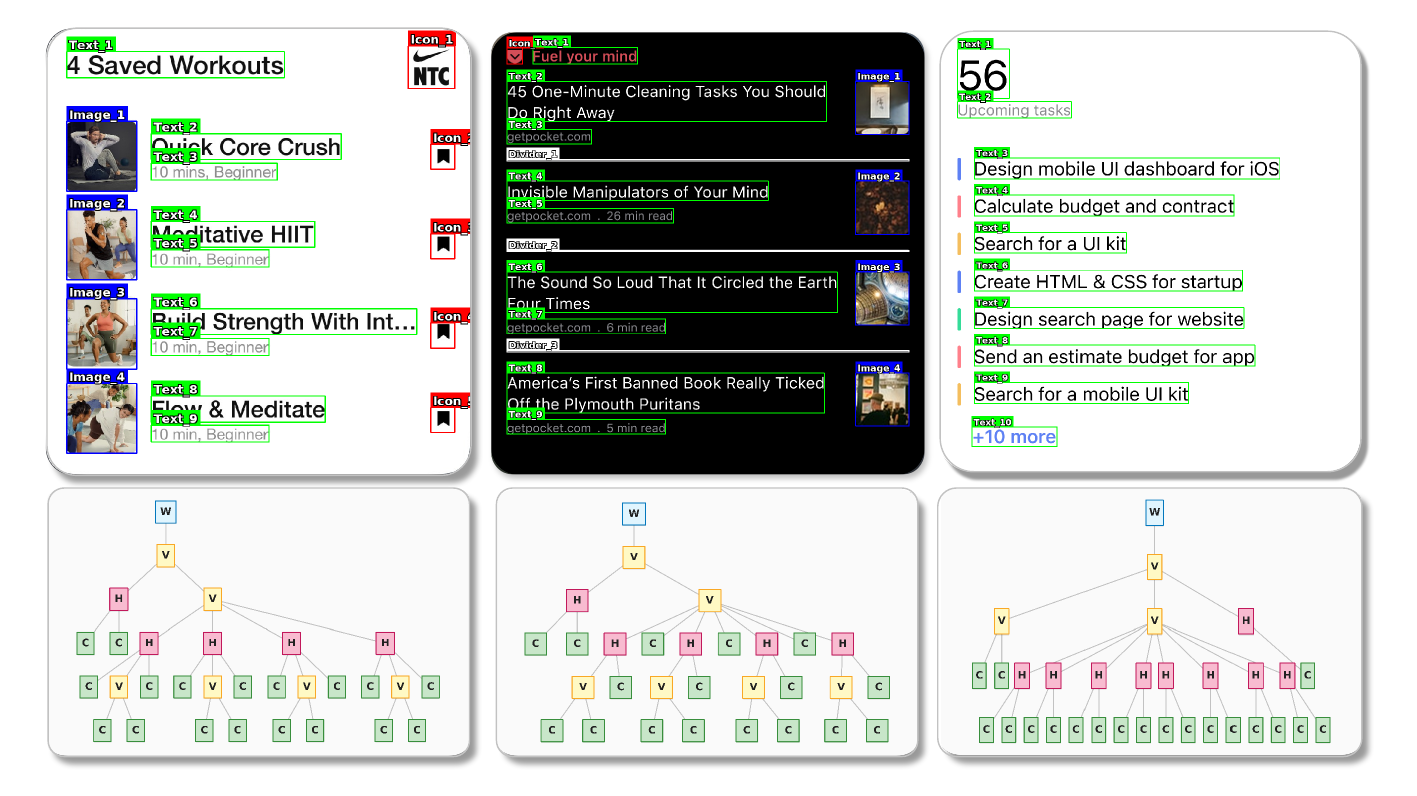}
  \caption{\textbf{Qualitative results of Qwen3-VL-8B fine-tuned on synthetic data from WidgetFactory, evaluated on real-world widget screenshots.} The first row shows predictions from the general grounding task, where the model detects and categorizes UI elements. The second row visualizes layout generation outputs as hierarchical trees, with \textbf{H} indicating horizontal containers, \textbf{V} indicating vertical containers, and \textbf{C} representing leaf-level UI components.}
  \label{fig:qwen3vl-results}
\end{figure}

\begin{table*}
  \tabcolsep3pt
  \centering
  \begin{tabular}{lccccccccccccc}
    \toprule
    Model & \textbf{Marg.} & \textbf{Cont.} & \textbf{Area} & \textbf{Text} & \textbf{Cont.} & \textbf{Loc.} & \textbf{Pale.} & \textbf{Vib.} & \textbf{Pola.} & \textbf{SIM} & \textbf{LP.$\downarrow$} & \textbf{CLIP} & \textbf{Geo}\\
    \cmidrule(r){2-4} \cmidrule(r){5-7} \cmidrule(r){8-10} \cmidrule(r){11-13} \cmidrule(r){14-14}
    Qwen8B & 49.82 & 52.38 & 50.51 & 27.69 & 54.38 & 29.19 & 23.96 & 22.78 & 36.24 & 0.422 & 0.523 & 0.689 & 27.37\\
    \rowcolor{gray!20} Finetune & 56.48 & 62.68 & 68.78 & 32.87 & 52.29 & 33.88 & 20.55 & 20.14 & 35.32 & 0.498 & 0.549 & 0.712 & 38.95\\
    \hline
    Ours& 66.16 & 59.29 & 72.63 & 66.05 & 58.89 & 60.46 & 40.28 & 37.82 & 53.88 & 0.691 & 0.362 & 0.768& 100\\
    \rowcolor{gray!20} Ours**& 68.78 & 62.29 & 75.81 & 68.95 & 62.98 & 62.43 & 52.61 & 46.13 & 59.51 & 0.709 & 0.343 &  0.789& 100\\
  \bottomrule
  \end{tabular}
  \caption{Fine-tuning improves the performance, but the training strategy is important to obtain optimal performance.}
  \label{tab:8b_finetune}
\end{table*}

\begin{table*}
  \tabcolsep3pt
  \centering
    \small
  \begin{tabular}{lccccccccccccc}
    \toprule
    Methods & \textbf{Marg.} & \textbf{Cont.} & \textbf{Area} & \textbf{Text} & \textbf{Cont.} & \textbf{Loc.} & \textbf{Pale.} & \textbf{Vib.} & \textbf{Pola.} & \textbf{SIM} & \textbf{LP.$\downarrow$} & \textbf{CLIP} & \textbf{Geo}\\
    \cmidrule(r){2-4} \cmidrule(r){5-7} \cmidrule(r){8-10} \cmidrule(r){11-13} \cmidrule(r){14-14}
    Qwen-32B & 51.63 & 53.59 & 64.02 & 30.12 & 51.84 & 31.79 & 24.58 & 23.42 & 36.66 & 0.436 & 0.530 & 0.786 & 28.73\\
    \midrule
    \rowcolor{gray!20} ScreenCoder & 22.19 & 11.46 & 31.04 & 13.77 & 25.35 & 24.66 & 32.15 & 33.62 & 3.99 & 0.101 & 0.512 & 0.582 & 44.56 \\
    + Finetune & 66.97 & 65.73 & 77.31 & 60.45 & 54.17 & 47.57 & 40.68 & 38.04 & 55.68 & 0.657 & 0.363& 0.801 & 60.87\\
    \midrule
    \rowcolor{gray!20} Ours-32B & 69.52 & 65.09 & 77.86 & 67.90 & 64.98 & 59.16 & 52.59 & 48.10 & 60.95 & 0.708 & 0.336 & 0.830 & 100.00\\
  \bottomrule
  \end{tabular}
  \caption{Finetuning the baselines and smaller model.}
\label{tab:tune_base}
\end{table*}

\subsection{Example Usage: Supervised Fine-Tuning of MLLMs Using Synthetic Widget Data}\label{grounding}

WidgetFactory provides a rich and compositional component library encompassing both atomic elements, such as Text and over 50{,}000 diverse icons collected from open-source icon repositories, as well as complex modules including LineChart, BarChart, and structured layout primitives. This allows for the generation of diverse and semantically rich widget instances at scale.

\noindent We leverage this capability to perform supervised fine-tuning (SFT) of multimodal large language models (MLLMs) such as Qwen3-VL, using synthetic widget images and metadata as supervision. Fine-tuned models exhibit improved performance in UI-specific tasks such as layout understanding, visual grounding of components, and structured code generation. The synthetic dataset also enables controlled curriculum design by varying widget complexity, layout depth, and component composition.

\noindent Beyond dataset scale and diversity, WidgetFactory also supports fine-grained control over task-specific supervision. We construct synthetic instruction–response pairs across four representative widget understanding tasks:

\begin{itemize}
    \item \textbf{Layout Generation}: Producing hierarchical code of the UI component tree with explicit spatial relationships.
    \item \textbf{General Grounding}: Detecting all visible UI elements and localizing them with bounding boxes.
    \item \textbf{Category Grounding}: Identifying all elements of a given type (e.g., Icon, Text) within the image.
    \item \textbf{Referring Expression Comprehension}: Given a bounding box, generating a textual description of the corresponding UI element, including type, label, and visual attributes.
\end{itemize}

\noindent All examples are synthesized through WidgetFactory's infrastructure, with control over layout structure, component density, and screen resolution. This enables systematic construction of training corpora with curriculum-aligned complexity, ensuring both breadth and depth of supervision.

\noindent We use the synthetic datasets generated by WidgetFactory to fine-tune Qwen3-VL-8B on four representative widget understanding tasks. This demonstrates the utility of WidgetFactory as a flexible engine for adapting MLLMs to UI understanding and structured prediction. Notably, the fine-tuned model generalizes effectively to real-world widget screenshots, producing precise grounding and layout predictions despite being trained exclusively on synthetic data. Fig.~\ref{fig:qwen3vl-results} shows qualitative results from the fine-tuned model on layout generation, grounding, and referring tasks.

\noindent We further integrate the fine-tuning into the whole pipeline. As shown in Table~\ref{tab:8b_finetune}, finetuning improves the 8B model. It can be viewed as a distillation process to distill the better coding knowledge from a larger model into smaller models~\cite{hamidi2024train,ye2026asmil}. However, fine-tuning only the Component Extraction as in Fig.~\ref{fig:qwen3vl-results} (Ours$^{**}$) while using the untuned for other modules performs best, showing that training strategy is important for future exploration. 
In Table~\ref{tab:tune_base}, we show that finetuning improves the ScreenCoder, but our baseline further benefits from our generated data. 

\paragraph{Future exploration} There are a few directions that can be explored in the future. The whole workflow of the baseline can be further improved with better perception analysis, e.g., better grounding module as shown in Sec.\ref{grounding}. In addition, the fixed number of component templates also limit the flexibility as there are always non-covered or newly emerged ones. Therefore, it is intuitive to evolve the component library to discover and embed the new contents~\cite{wu2023metagcd,wu2026talon}. Last but not least, a feedback loop can also be developed to provide additional verification to enhance the generated code via automatic visual inspection or directly using the proposed metrics. 

%% file: supplementary/sections/sec_3.tex
\lstdefinestyle{jsonstyle}{
  basicstyle=\ttfamily\footnotesize,
  breaklines=true,
  breakatwhitespace=false,
  columns=fullflexible,
  keepspaces=true,
  showstringspaces=false,
  upquote=true,
  prebreak=\mbox{},   
  postbreak=\mbox{},  
  literate=
    {"}{\char34}1
    {“}{\char34}1
    {”}{\char34}1
    {‘}{\char39}1
    {’}{\char39}1
}
\begin{figure*}[t]
   \centering
   \footnotesize
   \begin{tcolorbox}[colback=promptlightblue, colframe=darkblue, rounded corners, title=Component Extraction Prompt, fonttitle=\bfseries]
\texttt{You are an expert mobile-UI grounding assistant. OUTPUT JSON ONLY.}\\
\texttt{Return ONE array of objects:}\\
\texttt{\{"bbox":[x1,y1,x2,y2],"label":"\textless class\textgreater","description":"\textless tokens\textgreater"\}.}\\
\texttt{GOAL: HIGH RECALL. Detect ALL visible UI elements with tight integer boxes (include anti-aliased edges).}\\
\texttt{\#\# VISUAL CHARACTERISTICS (for grounding)}\\
\texttt{**Container** --- background panel/card grouping multiple elements; solid or blurred fill, often with rounded corners and padding.}\\
\texttt{**Icon** --- simple vector pictogram or symbol; flat color or simple gradient, usually clean strokes and geometric shapes.}\\
\texttt{**AppLogo** --- subset of Icon representing well-known global brands/products (Google, Chrome, Apple, Microsoft, Spotify, Twitter/X, Instagram, YouTube, GitHub, etc.); typically square with solid background and central mark.}\\
\texttt{**Text** --- any readable alphanumeric glyphs; labels, numbers, or titles. Ignore texts inside charts (except buttons).}\\
\texttt{**Image** --- photographic or illustrative content with complex textures or gradients.}\\
\texttt{**Button** --- filled rounded rect/pill/circle background (often colored or elevated) containing icon or text. Detect both Button and inner Icon/Text separately.}\\
\texttt{**Checkbox** --- circular or square tick control; may be hollow (unchecked) or filled with a checkmark (checked).}\\
\texttt{**Switch** --- horizontal pill track with circular thumb sliding left/right between off/on states.}\\
\texttt{**Slider** --- thin horizontal track with a larger circular thumb marking a value.}\\
\texttt{**Divider** --- 1--2 px straight line separator, horizontal or vertical.}\\
\texttt{**Indicator** --- narrow vertical colored bar or stripe for category/status.}\\
\texttt{\#\# CHART VISUAL CHARACTERISTICS}\\
\texttt{Each chart is ONE element.}\\
\texttt{Include title, axes, ticks, legends, and value labels inside its bbox.}\\
\texttt{Do **not** mark inner text, dividers, or gridlines separately.}\\
\texttt{**BarChart** --- vertical or horizontal rectangular bars of uniform width; single color per bar; used to compare categories.}\\
\texttt{**StackedBarChart** --- bars divided into multiple colored segments stacked together; shows composition or proportions.}\\
\texttt{**LineChart** --- one or more continuous lines connecting data points; often with axes and gridlines; shows time-based trends.}\\
\texttt{**PieChart** --- circle divided into wedge-shaped slices; donut variants count as PieChart; shows proportions of a whole.}\\
\texttt{**RadarChart** --- spider/web-style polygon chart with radial axes; grid of concentric shapes and lines connecting data points.}\\
\texttt{**ProgressBar** --- long thin bar partially filled to indicate completion percentage (horizontal or vertical).}\\
\texttt{**ProgressRing** --- circular ring partially filled with a colored arc; may contain icon, text, or be empty.}\\
\texttt{**Sparkline** --- tiny minimalist line chart showing a short trend, **without** axes, ticks, or grid.}\\
\texttt{\#\# DISAMBIGUATION}\\
\texttt{- Tiny line with no axes/grid $\rightarrow$ Sparkline.}\\
\texttt{- Bars with multiple colored segments $\rightarrow$ StackedBarChart.}\\
\texttt{- Donut circle $\rightarrow$ PieChart.}\\
\texttt{- Circular progress arc (partial fill) $\rightarrow$ ProgressRing.}\\
\texttt{- Choose the closest chart type by structure if uncertain.}\\
\texttt{\#\# FULL OUTPUT EXAMPLES (every element)}
\begin{lstlisting}[style=jsonstyle,aboveskip=0pt,belowskip=0pt]
[
  {"bbox":[12,16,280,180],"label":"Container","description":"shape:rounded,pad:16,bg:#1C1C1E,r:20"},
  {"bbox":[24,28,64,68],"label":"Icon","description":"type:heart,color:#FF3B30"},
  {"bbox":[72,28,112,68],"label":"AppLogo","description":"brand:Chrome,color:#FFFFFF"},
  {"bbox":[120,32,260,50],"label":"Text","description":"text:'Skills Assessment',color:#FFFFFF,weight:600"},
  {"bbox":[24,80,160,120],"label":"Image","description":"src:unsplash,shape:rect,w:136,h:40"},
  {"bbox":[24,140,84,200],"label":"Button","description":"shape:circle,bg:#007AFF,r:30,pad:10"},
]
\end{lstlisting}
   \end{tcolorbox}
   \caption{Prompt for extracting UI components from phone widget screenshots.}
   \label{fig:instruction_ui_grounding}
\end{figure*}

\begin{figure*}
   \centering
   \footnotesize
   \begin{tcolorbox}[colback=promptlightblue, colframe=darkblue, rounded corners, title=WidgetDSL Generation Prompt, fonttitle=\bfseries]
\texttt{\# Widget Specification Generation from Image}\\
\texttt{You are a VLM specialized in analyzing UI widget images and generating structured WidgetDSL in JSON format. Your task is to observe a widget image and output a complete, accurate WidgetDSL that can be compiled into a React component.}\\
\texttt{\#\#\# WidgetShell (Root Container)}\\
\texttt{Props: \textasciigrave{}backgroundColor\textasciigrave{} (hex), \textasciigrave{}borderRadius\textasciigrave{} (number), \textasciigrave{}padding\textasciigrave{} (number), \textasciigrave{}aspectRatio\textasciigrave{} (number)}\\
\texttt{- \textasciigrave{}aspectRatio\textasciigrave{}: [ASPECT\_RATIO]}\\
\texttt{\#\# Layout INFO}\\
\texttt{[LAYOUT\_INFO]}\\
\texttt{\#\# Detected Components (MUST USE)}\\
\texttt{[PRIMITIVE\_DEFINITIONS]}\\
\texttt{\#\# Optional Components (use if visible in image but not detected above)}\\
\texttt{[FALLBACK\_PRIMITIVES]}\\
\texttt{\#\#\# Color Palette}\\
\texttt{[COLOR\_PALETTE]}\\
\texttt{\#\#\# Graph}\\
\texttt{[GRAPH\_SPECS]}\\
\texttt{\#\# Layout System}\\
\texttt{All layouts use **flexbox containers**. There are two node types:}\\
\texttt{\#\#\# Container Node}
\begin{lstlisting}[style=jsonstyle,aboveskip=0pt,belowskip=0pt]
{"type": "container",
  "direction": "row" | "col",
  "gap": number,
  "flex": number | "none" | 0 | 1,
  "width": number | string (optional, for layout control),
  "height": number | string (optional, for layout control),
  "padding": number,
  "backgroundColor": "#hex",
  "borderRadius": number (optional),
  "children": [...]}
\end{lstlisting}
\texttt{**Layout Control**: Containers can have explicit \textasciigrave{}width\textasciigrave{} and \textasciigrave{}height\textasciigrave{} for precise sizing:}\\
\texttt{- Use numbers for fixed pixel values: \textasciigrave{}"width": 120\textasciigrave{} and strings for percentages: \textasciigrave{}"width": "50\%"\textasciigrave{}}\\
\texttt{- Combine with \textasciigrave{}flex\textasciigrave{} for responsive layouts}\\
\texttt{\#\#\# Leaf Node (Component)}
\begin{lstlisting}[style=jsonstyle,aboveskip=0pt,belowskip=0pt]
{"type": "leaf",
  "component": [AVAILABLE_COMPONENTS],
  "flex": number | "none" | 0 | 1,
  "width": number | string (optional, for layout control),
  "height": number | string (optional, for layout control),
  "props": { /* component-specific props */ },
  "content": "text content (for Text component only)"}
\end{lstlisting}
\texttt{**IMPORTANT**: For components like Image, Sparkline, and MapImage:}\\
\texttt{- Specify \textasciigrave{}width\textasciigrave{} and \textasciigrave{}height\textasciigrave{} at the **node level** (outside props)}\\
\texttt{- Example: \textasciigrave{}\{ "type": "leaf", "component": "Image", "width": 100, "height": 100, "props": \{ "src": "..." \} \}\textasciigrave{}}\\
\texttt{\#\# Output Format}\\
\texttt{Your output must be valid JSON following this structure:}
\begin{lstlisting}[style=jsonstyle,aboveskip=0pt,belowskip=0pt]
{"widget": {
    "backgroundColor": "#hex",
    "borderRadius": number,
    "padding": number,
    "aspectRatio": [ASPECT_RATIO],
    "root": {
      "type": "container",
      "direction": "col",
      "children": [...]}}}
\end{lstlisting}
   \end{tcolorbox}
   \caption{WidgetDSL generation instruction prompt to produce structured WidgetDSL from a phone widget screenshot.}
   \label{fig:widgetdsl_generation_prompt}
\end{figure*}

\begin{figure*}
   \centering
   \footnotesize
   \begin{tcolorbox}[colback=promptlightblue, colframe=darkblue, rounded corners, title=Primitive Generation - Icon, fonttitle=\bfseries]
\texttt{\#\#\# Icon}\\
\texttt{Props: \textasciigrave{}name\textasciigrave{} (string with prefix:Name), \textasciigrave{}size\textasciigrave{} (number), \textasciigrave{}color\textasciigrave{} (hex)}\\
\texttt{- **IMPORTANT**: Must use \textasciigrave{}"prefix:ComponentName"\textasciigrave{} format (e.g., \textasciigrave{}"sf:SfBoltFill"\textasciigrave{}, \textasciigrave{}"lu:LuHeart"\textasciigrave{})}\\
\texttt{- Prefixes: \textasciigrave{}ai\textasciigrave{}, \textasciigrave{}bi\textasciigrave{}, \textasciigrave{}bs\textasciigrave{}, \textasciigrave{}cg\textasciigrave{}, \textasciigrave{}ci\textasciigrave{}, \textasciigrave{}di\textasciigrave{}, \textasciigrave{}fa\textasciigrave{}, \textasciigrave{}fa6\textasciigrave{}, \textasciigrave{}fc\textasciigrave{}, \textasciigrave{}fi\textasciigrave{}, \textasciigrave{}gi\textasciigrave{}, \textasciigrave{}go\textasciigrave{}, \textasciigrave{}gr\textasciigrave{}, \textasciigrave{}hi\textasciigrave{}, \textasciigrave{}hi2\textasciigrave{}, \textasciigrave{}im\textasciigrave{}, \textasciigrave{}io\textasciigrave{}, \textasciigrave{}io5\textasciigrave{}, \textasciigrave{}lia\textasciigrave{}, \textasciigrave{}lu\textasciigrave{}, \textasciigrave{}md\textasciigrave{}, \textasciigrave{}pi\textasciigrave{}, \textasciigrave{}ri\textasciigrave{}, \textasciigrave{}rx\textasciigrave{}, \textasciigrave{}sf\textasciigrave{}, \textasciigrave{}si\textasciigrave{}, \textasciigrave{}sl\textasciigrave{}, \textasciigrave{}tb\textasciigrave{}, \textasciigrave{}tfi\textasciigrave{}, \textasciigrave{}ti\textasciigrave{}, \textasciigrave{}vsc\textasciigrave{}, \textasciigrave{}wi\textasciigrave{}}\\
\texttt{- Available icon names: [AVAILABLE\_ICON\_NAMES]}\\
\texttt{- Example:}
\begin{lstlisting}[style=jsonstyle,aboveskip=0pt,belowskip=0pt]
{
  "type": "leaf",
  "component": "Icon",
  "props": {
    "name": "sf:SfHeart",
    "size": 24,
    "color": "#FF0000"
  }
}

\end{lstlisting}
   \end{tcolorbox}
   \caption{Primitive generation instruction for icons.}
   \label{fig:icon_primitive_definition}
\end{figure*}

\begin{figure*}
   \centering
   \footnotesize
   \begin{tcolorbox}[colback=promptlightblue, colframe=darkblue, rounded corners, title=Primitive Generation - AppLogo, fonttitle=\bfseries]
\texttt{\#\#\# AppLogo}\\
\texttt{Props: \textasciigrave{}icon\textasciigrave{} (string, optional), \textasciigrave{}name\textasciigrave{} (string), \textasciigrave{}size\textasciigrave{} (number), \textasciigrave{}backgroundColor\textasciigrave{} (hex, optional)}\\
\texttt{- **IMPORTANT**: If \textasciigrave{}icon\textasciigrave{} prop is provided, use it (e.g., \textasciigrave{}"si:SiGoogle"\textasciigrave{}, \textasciigrave{}"si:SiSpotify"\textasciigrave{})}\\
\texttt{- Otherwise displays first letter of \textasciigrave{}name\textasciigrave{} with rounded square background}\\
\texttt{- Border radius auto-calculated (22\% of size)}\\
\texttt{- Available applogo names (brand/app icons): [AVAILABLE\_APPLOGO\_NAMES]}
\begin{lstlisting}[style=jsonstyle,aboveskip=0pt,belowskip=0pt]
{
  "type": "leaf",
  "component": "AppLogo",
  "props": {
    "name": "Music",
    "size": 40,
    "backgroundColor": "#FF3B30"
  }
}

\end{lstlisting}
   \end{tcolorbox}
   \caption{Primitive generation instruction for app logos.}
   \label{fig:applogo_primitive_definition}
\end{figure*}

\begin{figure*}
    \centering
    \footnotesize

    \begin{tcolorbox}[
        colback=promptlightblue,
        colframe=darkblue,
        rounded corners,
        title=Primitive Generation - Button,
        fonttitle=\bfseries
    ]
\texttt{\#\#\# Button}\\
\texttt{Props: \textasciigrave{}icon\textasciigrave{} (string), \textasciigrave{}backgroundColor\textasciigrave{} (hex), \textasciigrave{}color\textasciigrave{} (hex), \textasciigrave{}borderRadius\textasciigrave{} (number),}\\
\texttt{\phantom{Props: } \textasciigrave{}fontSize\textasciigrave{} (number), \textasciigrave{}fontWeight\textasciigrave{} (number), \textasciigrave{}padding\textasciigrave{} (number), \textasciigrave{}content\textasciigrave{} (text)}\\
\texttt{Node properties: \textasciigrave{}width\textasciigrave{} (number), \textasciigrave{}height\textasciigrave{} (number)}\\
\texttt{- \textbf{RARE in widgets} — only use when a clear button with background/padding exists}\\
\texttt{- Contains either icon OR text (never both)}\\
\texttt{- Circular button: set \textasciigrave{}borderRadius = size/2\textasciigrave{}}\\
\texttt{- Example:}
\begin{lstlisting}[style=jsonstyle,aboveskip=0pt,belowskip=0pt]
{
  "type": "leaf",
  "component": "Button",
  "props": {
    "icon": "sf:SfPlus",
    "backgroundColor": "#007AFF",
    "color": "#fff",
    "borderRadius": 12,
    "padding": 12
  }
}

\end{lstlisting}
    \end{tcolorbox}
    \caption{Primitive generation instructions for buttons.}
    \label{fig:button_primitive_definition}
\end{figure*}

\begin{figure*}
   \centering
   \footnotesize
   \begin{tcolorbox}[colback=promptlightblue, colframe=darkblue, rounded corners, title=WidgetDSL Specification Prompt - Barchart, fonttitle=\bfseries]
\texttt{Generate a WidgetDSL specification for a BarChart component in this image.}\\
\texttt{Focus on extracting these elements:}\\
\texttt{\#\# Chart Identification}\\
\texttt{- **Title**: Extract title text (set showTitle: false if none)}\\
\texttt{- **Orientation**: "vertical" (bars go up) or "horizontal" (bars go right)}\\
\texttt{- **Data Series**: Single series or multiple series (grouped bars)}\\
\texttt{- **Category Labels**: List ALL labels in exact order}\\
\texttt{\#\# Data Extraction}\\
\texttt{**For single series**: Array of numbers [10, 20, 15, 30]}\\
\texttt{**For multiple series**: 2D array [[series1], [series2], ...]}\\
\texttt{- Extract exact bar heights/lengths as values}\\
\texttt{- Note any bars with zero values}\\
\texttt{- For grouped bars, maintain series order from legend}\\
\texttt{\#\# Axis Configuration - IMPORTANT}\\
\texttt{**The component automatically calculates axis ranges:**}\\
\texttt{- **min**: Defaults to 0 for positive values, uses smart rounding for negative values}\\
\texttt{- **max**: Automatically calculated using smart rounding}\\
\texttt{\ \ \ - Applies magnitude-aware rounding (e.g., 65 $\rightarrow$ 80, 847 $\rightarrow$ 1000, 23 $\rightarrow$ 30)}\\
\texttt{\ \ \ - Adds 10\% padding for better visualization}\\
\texttt{\ \ \ - Override by setting explicit \textasciigrave{}max\textasciigrave{} value if exact range needed}\\
\texttt{**Only specify axis values when:**}\\
\texttt{- Image shows explicit max value (e.g., "0-100" scale)}\\
\texttt{- Exact intervals are visible (e.g., grid lines every 10 units)}\\
\texttt{- Otherwise, omit \textasciigrave{}max\textasciigrave{} and \textasciigrave{}interval\textasciigrave{} to use automatic calculation}\\
\texttt{\#\# Visual Styling}\\
\texttt{- **Colors**: For single series, one color; for multiple series, array of colors}\\
\texttt{- **Background**: Chart background color}\\
\texttt{- **Theme**: "light" or "dark"}\\
\texttt{- **Grid lines**: Color, style (solid/dashed), visibility}\\
\texttt{- **Bar styling**: Width, border radius, spacing between bars}\\
\texttt{- **Value labels**: Whether values are displayed on/above bars}\\
\texttt{\#\# Axis Label Customization (Advanced)}\\
\texttt{- **Label colors**: \textasciigrave{}xAxisLabelColor\textasciigrave{}, \textasciigrave{}yAxisLabelColor\textasciigrave{} - Custom colors for axis labels}\\
\texttt{- **Label sizes**: \textasciigrave{}xAxisLabelFontSize\textasciigrave{}, \textasciigrave{}yAxisLabelFontSize\textasciigrave{} - Font size in pixels (default: 11)}\\
\texttt{- **Label rotation**: \textasciigrave{}xAxisLabelRotate\textasciigrave{}, \textasciigrave{}yAxisLabelRotate\textasciigrave{} - Rotation angle in degrees (0 = horizontal)}\\
\texttt{- **Label suffixes**: \textasciigrave{}xAxisLabelSuffix\textasciigrave{}, \textasciigrave{}yAxisLabelSuffix\textasciigrave{} - Add suffix to axis values (e.g., "m" for minutes, "\%" for percentages)}\\
\texttt{- **Custom formatters**: For special number formatting (use sparingly)}\\
   \end{tcolorbox}
   \caption{Prompt for generating a structured WidgetDSL BarChart specification from a widget screenshot.}
   \label{fig:barchart_widgetdsl_prompt}
\end{figure*}

\begin{figure*}
   \centering
   \footnotesize
   \begin{tcolorbox}[colback=promptlightblue, colframe=darkblue, rounded corners, title=Widget-to-HTML Baseline Prompt, fonttitle=\bfseries]
   \texttt{Given a phone widget screenshot, generate ONE single self-contained HTML file that reproduces the widget UI.
}\\
   \texttt{Rules:}\\
   \texttt{- Output ONLY HTML code. No explanations, no comments.}\\
   \texttt{- Begin exactly with: \textless html lang="en"\textgreater\ and end exactly with: \textless/html\textgreater.}\\
   \texttt{- Place all widget content inside exactly one container: \textless div class="widget"\textgreater\ ... \textless/div\textgreater\ in the \textless body\textgreater.}\\
   \texttt{The widget must maintain the screenshot's aspect ratio ($\approx$ 1.00:1) as closely as possible.}

   \end{tcolorbox}
   \caption{Baseline widget-to-HTML instruction prompt to generate a single self-contained HTML file from a phone widget screenshot.}
   \label{fig:instruction_concept_dictionary}
\end{figure*}

%% file: supplementary/sections/component.tex
\begin{figure*}
    \centering
    \includegraphics[width=0.9\linewidth]{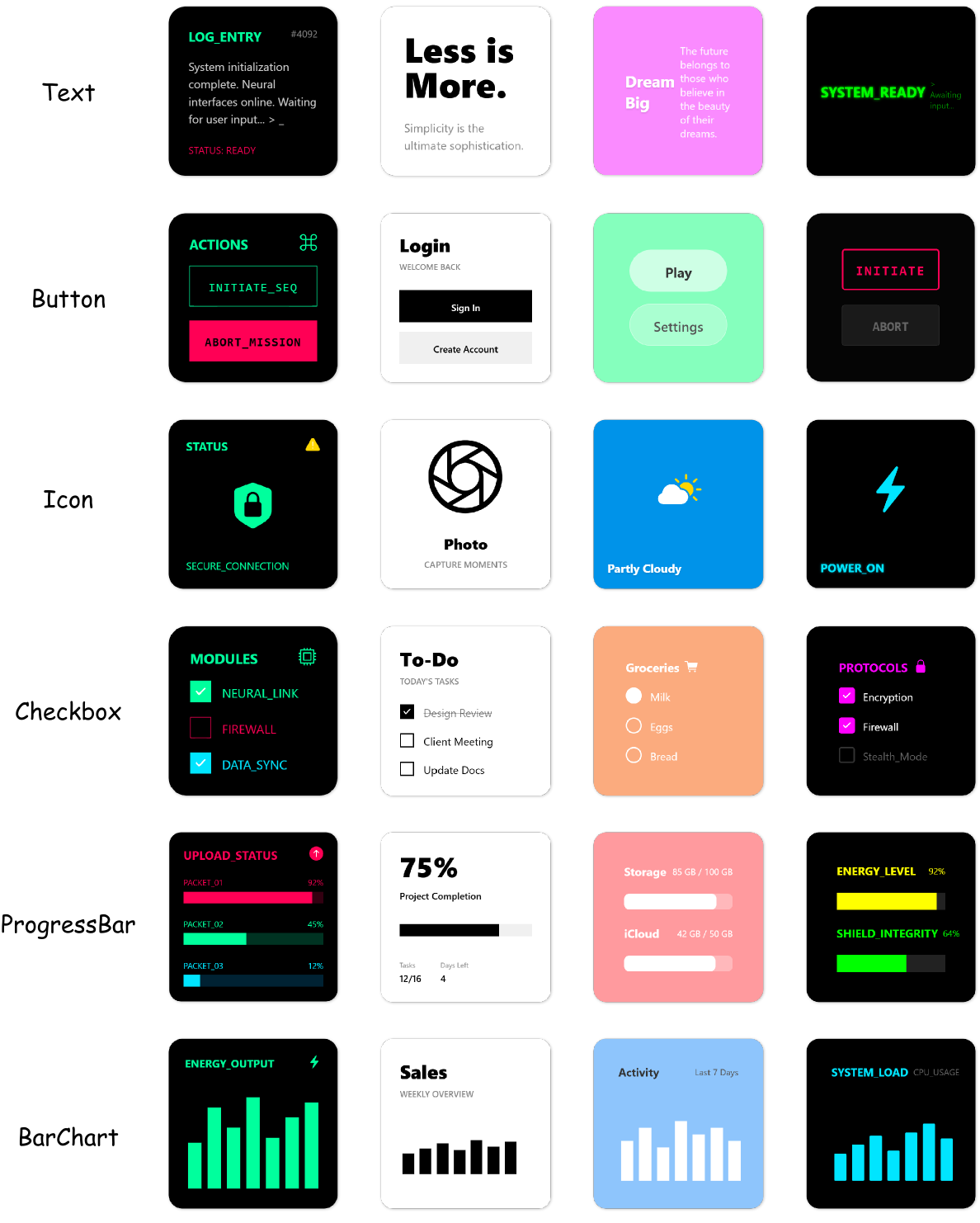}
    \label{fig:supp_component_more}
    \caption{Illustration of rendered images from the component templates.}
\end{figure*}

\pdfcompresslevel=0
\pdfobjcompresslevel=0

\begin{figure*}
    \centering
    \includegraphics[width=0.9\linewidth]{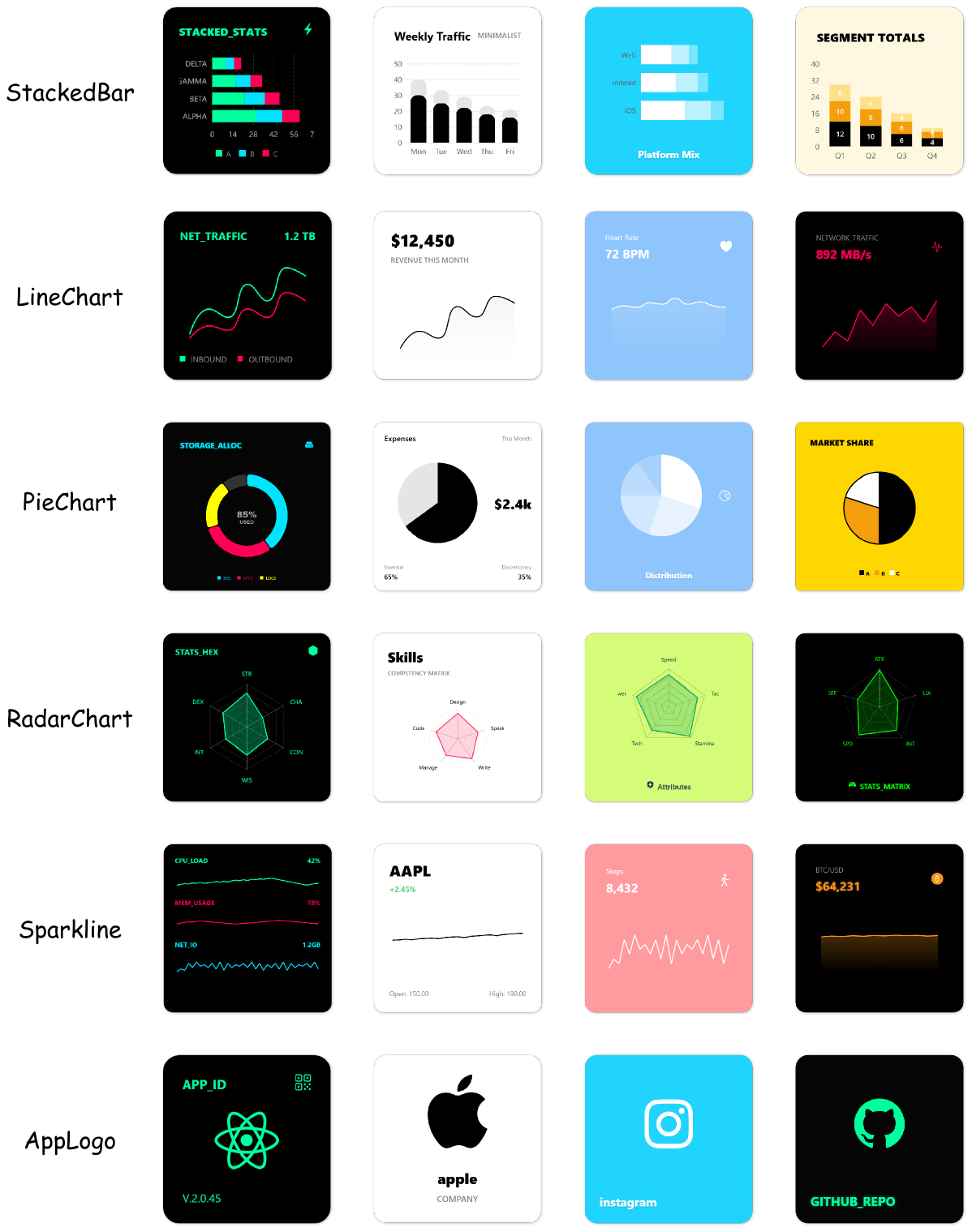}
    \caption{Illustration of rendered images from the component templates.}
    \label{fig:supp_component_more_1}
\end{figure*}